\documentclass[11pt]{article}

\usepackage[T1]{fontenc}
\usepackage[a4paper,margin=1.1in]{geometry}
\usepackage{calc}
\usepackage[authoryear,round]{natbib}
\usepackage{amsmath,amssymb,amsthm,mathtools,bm}
\usepackage[expansion=false]{microtype}
\usepackage{graphicx}
\usepackage{subcaption}
\usepackage{booktabs,multirow,array,adjustbox}
\usepackage{algorithm}
\usepackage{algorithmic}
\usepackage{float}

\usepackage{hyperref}
\usepackage{cleveref}

\hypersetup{colorlinks=true,linkcolor=blue,citecolor=blue,urlcolor=blue}

\newcommand{\R}{\mathbb{R}}
\newcommand{\de}{\mathrm{d}}

\newcommand{\tr}{\operatorname{tr}}

\DeclareMathOperator*{\argmax}{arg\,max}

\newcommand{\pinnfig}[2]{%
  \IfFileExists{#1}{\includegraphics[width=#2]{#1}}{%
    \fbox{\begin{minipage}[c][0.22\textheight][c]{#2}\centering\ttfamily\small
      [figure not found]\\[2pt]\detokenize{#1}%
    \end{minipage}}%
  }%
}

\newtheorem{theorem}{Theorem}[section]
\newtheorem{proposition}[theorem]{Proposition}
\newtheorem{lemma}[theorem]{Lemma}

\newtheorem{definition}[theorem]{Definition}
\newtheorem{assumption}[theorem]{Assumption}
\newtheorem{remark}[theorem]{Remark}

\begin{document}

\title{Physics-Informed Policy Iteration for High-Dimensional
Hamilton--Jacobi--Bellman Equations: Interior Error Bounds
without Boundary Data}

\author{%
Yeongjong Kim\thanks{These authors contributed equally to this work.}$^{,1}$
\and
Minseok Kim$^{*,2}$
\and
Yeoneung Kim\thanks{Corresponding authors:
\texttt{yeoneung@yonsei.ac.kr}, \texttt{namkyeong@gachon.ac.kr}.}$^{,3}$
\and
Namkyeong Cho$^{\dagger,4}$
}

\date{}
\maketitle

\begin{center}
\small
$^{1}$Department of Mathematics, POSTECH, 77 Cheongam-ro, Nam-gu,
Pohang 37673, Republic of Korea\\
$^{2}$Department of Applied Artificial Intelligence, Seoul National
University of Science and Technology, 232 Gongneung-ro, Nowon-gu,
Seoul 01811, Republic of Korea\\
$^{3}$Department of Industrial Engineering, Yonsei University,
50 Yonsei-ro, Seodaemun-gu, Seoul 03722, Republic of Korea\\
$^{4}$Gachon University, 1342 Seongnam-daero, Sujeong-gu,
Seongnam 13120, Gyeonggi-do, Republic of Korea
\end{center}

\begin{abstract}
We develop a physics-informed policy-iteration method for stationary
second-order Hamilton--Jacobi--Bellman equations arising in continuous-time
stochastic control. Each policy-evaluation step is a linear elliptic PDE and is
approximated by a mesh-free neural residual solver; policy improvement is then
performed pointwise from the surrogate gradient. The analysis addresses
bounded-domain training without prescribed boundary data. We prove
well-posedness of the PDE-defined evaluation for every Borel Markov policy,
establish Lipschitz stability of the greedy map on bounded gradient ranges,
and derive an exponential attenuation estimate for unresolved boundary
information. These ingredients yield a closed finite-step interior error bound
whose floors are determined by a continuous $L^p$ residual, with a finite
exponent $p>d$, and an attenuated amplitude term. The experiments measure the
quantities in the estimate. On a linear--quadratic testbed with an exact
reference, the gradient-error floor scales nearly linearly with a
fresh-sample $L^p$ residual estimate, and finite-difference probes identify
when a training buffer is beneficial. At matched architecture and budget,
linear fixed-policy training becomes markedly more reliable than direct
minimization of the nonlinear HJB residual as the tested problems become more
difficult. The method also produces effective feedback on an inverted
pendulum, a planar quadrotor, and a 100-dimensional posterior-seeking
problem. These nonlinear tests expose two limitations not visible from
sampled residuals alone: the collocation distribution may miss the region
visited by the learned closed loop, and evaluation without boundary data may
leave undetermined the gradient component on which the greedy update depends.
On-policy collocation, rollout-anchored evaluation, and rollout-based
stopping provide effective model-only safeguards.
\end{abstract}

\noindent\textbf{Keywords:} stochastic optimal control; Hamilton--Jacobi--Bellman
equation; policy iteration; physics-informed neural networks; elliptic partial
differential equations; high-dimensional PDEs; scientific machine learning

\vspace{1em}


\section{Introduction}

Infinite-horizon stochastic optimal control leads to nonlinear
Hamilton--Jacobi--Bellman (HJB) equations. Grid-based solvers are reliable in
low dimensions but become impractical as the state dimension grows. Howard's
policy iteration (PI)~\citep{howard1960dynamic,puterman1979convergence,Puterman1981}
replaces the nonlinear HJB equation by a sequence of linear fixed-policy
evaluation equations followed by pointwise greedy updates. Its computational
bottleneck is therefore the repeated solution of high-dimensional linear
elliptic PDEs.

We study a mesh-free realization in which each fixed-policy PDE is
approximated by a physics-informed neural network (PINN). The resulting method,
PINN-PI, retains the equation-level structure of policy iteration instead of
minimizing the fully nonlinear HJB residual directly. This distinction is
important because policy improvement depends on the value gradient: a small
value error, or even a small empirical residual, need not yield an accurate
control law.

The central analytical issue comes from the computational domain. Although the
control problem is posed on $\R^d$, neural training is performed on a bounded
set, often without reliable boundary values. A residual-only estimate is
impossible there because the homogeneous fixed-policy equation admits nonzero
solutions. We therefore quantify how unresolved boundary information enters
the interior. For the coercive fixed-policy operator, we prove exponential
attenuation with the distance to the computational boundary. Combining this
estimate with local elliptic regularity, an a priori network-amplitude bound,
and Lipschitz stability of the greedy map yields a closed finite-step interior
error recursion. The bound separates the neural evaluation residual, the
attenuated outer amplitude, propagation through policy improvement, and the
convergence error of ideal PI.

The numerical study tests these mechanisms rather than aggregate performance
alone. A linear--quadratic benchmark provides exact policy
evaluations and an exact optimal value, allowing direct measurement of the
attenuation rate, gradient-error recursion, residual floor, and dimension
scaling. We then study an inverted pendulum, a planar quadrotor, and a
100-dimensional controlled Langevin problem. These examples show both the
advantage of fixed-policy training over direct nonlinear HJB training and two
failure modes that sampled residuals may miss: the learned closed loop can
visit regions poorly represented by the collocation distribution, and
evaluation without boundary data can leave the surrogate free in exactly the
gradient component on which a saturated greedy update depends. We test three model-only
safeguards: trajectory-adapted collocation, policy evaluation anchored at
rollout-estimated values, and rollout-based
stopping. Comparisons with model-free reinforcement learning are outside our
scope because PINN-PI assumes access to the dynamics.

\paragraph{Contributions}
\begin{itemize}
  \item We formulate PINN-based policy iteration for stationary second-order
  HJB equations and prove that every Borel Markov policy generated by the ideal
  or neural update has a unique PDE-defined evaluation in a Lyapunov growth
  class.
  \item We establish a gradient-to-policy Lipschitz estimate and an
  exponential interior attenuation estimate for unresolved boundary data.
  \item On nested domains, we derive a closed finite-step error bound with
  geometric iteration terms and floors determined by a continuous $L^p$
  residual, with $p>d$, and attenuated outer amplitude, under a sufficient
  feedback-contraction condition.
  \item We measure the individual terms of the error estimate against an
  exact reference, isolate the effect of the fixed-policy structure at
  matched budget, and identify and mitigate the collocation--closed-loop
  mismatch that the nonlinear benchmarks expose.
\end{itemize}

\section{Related Work}

\paragraph{Policy iteration for controlled diffusions}
Policy iteration originates with \citet{howard1960dynamic} and was connected
to Newton--Kantorovich methods and convergence rates by
\citet{puterman1979convergence}. Continuous-time controlled diffusions were
studied by \citet{Puterman1981}. More recent analyses address weak,
probabilistic, or viscosity settings
\citep{jacka2017policy,KerimkulovSiskaSzpruch2020,tang2025policy}, as well as
entropy-regularized and exploratory control
\citep{tran2025policy,huang2025convergence}. These works concern the ideal
policy-improvement sequence. Here its convergence is isolated as an assumption,
and the analysis closes the additional error introduced by neural
fixed-policy solves on bounded domains.

\paragraph{Approximate policy iteration and neural PDE solvers}
Classical approximate policy- and value-iteration bounds propagate evaluation
errors through Bellman or policy-improvement operators
\citep{munos2003error,farahmand2010error,dai2018sbeed}. They do not directly
address two PDE-specific issues considered here: the greedy control depends on
$\nabla v$, and a fixed-policy residual on a bounded domain does not control
the solution without outer information. PINNs provide mesh-free residual
approximations for PDEs~\citep{raissi2019physics}, and related mesh-free
formulations solve high-dimensional PDEs by deep Galerkin or BSDE-based
methods~\citep{sirignano2018dgm,han2018solving}; neural PI variants have been
developed mainly for deterministic control and first-order HJB equations
\citep{meng2024physics}. Related directions include nonconvex HJI and viscous Hamilton--Jacobi
equations, viscosity treatments of stationary discounted HJB equations,
operator-learning formulations, and physics-informed or model-based methods
for exploratory and reinforcement-learning formulations
\citep{yang2025solving,guo2026policy,cho2026policy,lee2025hamilton,%
kim2026physics,ramesh2023physics,mukherjee2023bridging}.
Our contribution is a second-order stochastic-control analysis that combines
interior elliptic stability, gradient-to-policy error propagation, and
bounded-domain training without prescribed boundary values.

\section{Infinite-horizon stochastic optimal control}

Let $(W_t)_{t\ge0}$ be a $d$-dimensional Brownian motion on a filtered
probability space $(\Xi,\mathcal F,(\mathcal F_t)_{t\ge0},\mathbb P)$. A
bounded progressively measurable control process
$a_t\in A\subset\mathbb R^m$ drives the controlled diffusion
\begin{equation}\label{eq:SDE}
  \mathrm dX_t
  =
  b(X_t,a_t)\,\mathrm dt
  +
  \sigma\,\mathrm dW_t,
  \qquad
  X_0=x\in\mathbb R^d,
\end{equation}
where $\sigma$ is a constant matrix. The goal is to maximize the
infinite-horizon discounted reward
\begin{equation}\label{eq:cost}
  J(x,a)
  =
  \mathbb E_x\left[
      \int_0^\infty e^{-\lambda t}
      L(X_t,a_t)\,\mathrm dt
  \right],
  \qquad
  \lambda>0 .
\end{equation}
The value function is defined by
\[
V(x):=\sup_a J(x,a).
\]
Under suitable regularity and growth assumptions on $b$ and $L$, the value
function is formally characterized as a solution of the HJB equation
\begin{equation}\label{eq:v}
\lambda V
-
\frac12\operatorname{tr}(\sigma\sigma^\top D_{xx}^2 V)
-
\sup_{a\in A}
\left\{
b(x,a)\cdot\nabla_x V
+
L(x,a)
\right\}
=
0
\qquad \text{in } \mathbb R^d .
\end{equation}
Formally, an optimal Markov policy is obtained by the pointwise greedy rule
\[
a^\ast(x)
\in
\argmax_{a\in A}
\left\{
b(x,a)\cdot\nabla_x V(x)
+
L(x,a)
\right\}.
\]
This greedy maximization recurs at every policy-improvement step. For a
gradient value $z\in\mathbb R^d$, define the set-valued greedy map
\begin{equation}\label{eq:greedy_map}
\mathcal A^\ast(x,z):=\operatorname*{Argmax}_{a\in A}\{b(x,a)\cdot z+L(x,a)\},
\end{equation}
which is nonempty and compact by (A1)--(A3) and admits Borel measurable
selectors by the measurable maximum theorem. Strong concavity of $L(x,\cdot)$
alone does not guarantee single-valuedness when $b$ is nonlinear in the
control: on gradient ranges $|z|\le M$ satisfying the concavity-dominance
condition $\mu_a>L_{ab}M$ imposed later
(Assumption~\ref{ass:concavity_dominance}), and automatically when $b$ is
affine in the control, the set $\mathcal A^\ast(x,z)$ is a singleton, and we
then write $a^\ast(x,z)$ for its unique element. Generated policies are
always understood as Borel measurable selectors of $\mathcal A^\ast$; the
single-valued notation $a^\ast$ is used only where the dominance condition
applies.

\begin{remark}[PDE-level scope of the analysis]
\label{rem:pde_scope}
Throughout the paper, the optimal value function is understood as a solution
of the HJB equation~\eqref{eq:v}, and, for a Markov policy $\alpha$, the
\emph{exact policy evaluation} $v^\alpha$ denotes the unique strong solution
of the linear fixed-policy equation provided by Theorem~\ref{thm:wellposed}
below. The stochastic control problem~\eqref{eq:SDE}--\eqref{eq:cost}
serves as the motivation for these equations. The identification of
$v^\alpha$ with the probabilistic payoff $J(\cdot,\alpha)$ of an arbitrary
Borel measurable Markov policy requires additional stochastic well-posedness
and representation arguments (strong solvability of the state equation
under measurable drift, an It\^o--Krylov formula for
$W^{2,p}_{\mathrm{loc}}$ functions, and a verification theorem), which
hold under classical additional conditions, for instance for bounded
measurable drift with nondegenerate constant diffusion
\citep{veretennikov1981strong,krylov1980controlled}, but are neither needed
for, nor claimed by, the error analysis of this paper. All statements below
are therefore statements about the PDE-defined quantities.
\end{remark}

\paragraph{Notation}
We use $|\cdot|$ for the Euclidean norm and standard $L^p$ and $W^{k,p}$
notation. Throughout the analysis, a finite exponent $p>d$ is fixed. All
strong-solution, residual, and interior regularity statements below refer to
this same exponent. The restriction $p>d$ is used only to obtain the embedding
$W^{2,p}_{\mathrm{loc}}\hookrightarrow C^{1,\alpha}_{\mathrm{loc}}$, which is
needed because policy improvement depends pointwise on the value gradient.
Throughout, $C$ denotes a positive constant that may change from line to
line; the quantities it depends on are displayed wherever that dependence
matters.

\begin{assumption}[Standing assumptions]
\label{ass:main}
We impose the following assumptions throughout the paper.

\begin{enumerate}

\item[(A1)] \textbf{Control set.}
The action set $A\subset\mathbb R^m$ is nonempty, compact, and convex.

\item[(A2)] \textbf{Dynamics regularity.}
The drift $b:\mathbb R^d\times A\to\mathbb R^d$ is Borel measurable in $x$
and continuously differentiable in $a$. Moreover, it satisfies the following
conditions.

\begin{itemize}

\item \textbf{Uniform local boundedness in $x$:}
for every bounded domain $K\subset\mathbb R^d$,
\[
\Big\|\sup_{a\in A}|b(\cdot,a)|\Big\|_{L^\infty(K)}<\infty .
\]

\item \textbf{Lipschitz continuity in control:}
there exists $B_a>0$ such that
\[
|b(x,a)-b(x,a')|
\le
B_a|a-a'|,
\qquad
x\in\mathbb R^d,\quad a,a'\in A .
\]

\item \textbf{Smooth dependence on control:}
the partial derivative $\partial_a b(x,a)$ exists and satisfies
\[
\sup_{x\in\mathbb R^d,\;a\in A}
\|\partial_a b(x,a)\|
\le
B_a ,
\]
and there exists $L_{ab}\ge0$ such that
\[
\|\partial_a b(x,a)-\partial_a b(x,a')\|
\le
L_{ab}|a-a'|,
\qquad
x\in\mathbb R^d,\quad a,a'\in A .
\]

\end{itemize}

\item[(A3)] \textbf{Reward regularity and concavity.}
The running reward $L:\mathbb R^d\times A\to\mathbb R$ is Borel measurable in
$x$ and continuously differentiable in $a$. Moreover, it satisfies the following
conditions.

\begin{itemize}

\item \textbf{Uniform local boundedness in $x$:}
for every bounded domain $K\subset\mathbb R^d$,
\[
\Big\|\sup_{a\in A}|L(\cdot,a)|\Big\|_{L^\infty(K)}<\infty .
\]

\item \textbf{Lipschitz continuity in control:}
there exists $L_a>0$ such that
\[
|L(x,a)-L(x,a')|
\le
L_a|a-a'|,
\qquad
x\in\mathbb R^d,\quad a,a'\in A .
\]

\item \textbf{Strong concavity in control:}
for each $x\in\mathbb R^d$, the map $a\mapsto L(x,a)$ is
$\mu_a$-strongly concave on $A$.

\end{itemize}

\item[(A4)] \textbf{Diffusion.}
The diffusion matrix $\sigma\in\mathbb R^{d\times d}$ is constant and uniformly
elliptic. That is, there exist constants $0<\nu\le\Lambda<\infty$ such that
\[
\nu|\xi|^2
\le
\xi^\top\sigma\sigma^\top\xi
\le
\Lambda|\xi|^2,
\qquad
\xi\in\mathbb R^d .
\]

\item[(A5)] \textbf{Lyapunov growth condition.}
There exist a function $\Psi\in C^2(\mathbb R^d)$ with $\Psi\ge1$ and constants
$C_L>0$ and $C_\Psi\ge0$ such that $\lambda>C_\Psi$ and
\[
\sup_{a\in A}|L(x,a)|
\le
C_L\Psi(x),
\qquad
x\in\mathbb R^d,
\]
and
\[
\sup_{a\in A}
\left\{
\frac12\operatorname{tr}(\sigma\sigma^\top D_{xx}^2\Psi(x))
+
b(x,a)\cdot\nabla_x\Psi(x)
\right\}
\le
C_\Psi\Psi(x),
\qquad
x\in\mathbb R^d .
\]
In addition, there exist a second weight $\Phi\in C^2(\mathbb R^d)$ with
$\Phi\ge1$ and a constant $C_\Phi\ge0$ such that $\lambda>C_\Phi$,
\[
\frac{\Psi(x)}{\Phi(x)}\to0
\qquad\text{as }|x|\to\infty,
\]
and
\[
\sup_{a\in A}
\left\{
\frac12\operatorname{tr}(\sigma\sigma^\top D_{xx}^2\Phi(x))
+
b(x,a)\cdot\nabla_x\Phi(x)
\right\}
\le
C_\Phi\Phi(x),
\qquad
x\in\mathbb R^d .
\]

\end{enumerate}
\end{assumption}

Assumptions~\ref{ass:main}(A1)--(A4) provide the local structure used in the
greedy update and elliptic estimates. Compactness of $A$ gives existence of
greedy maximizers; uniqueness on bounded gradient ranges additionally requires
the concavity-dominance condition introduced in
Assumption~\ref{ass:concavity_dominance}, which is automatic for
control-affine dynamics. The local coefficient bounds are uniform over Borel
selectors. Assumption~\ref{ass:main}(A5) supplies two Lyapunov weights: $\Psi$
controls the data and policy evaluations, while the faster weight $\Phi$
enables comparison at infinity. In dissipative models, one may often take
$\Psi(x)=(1+|x|^2)^r$ and $\Phi(x)=(1+|x|^2)^s$ with $s>r$, provided the
generator inequalities and discount conditions hold uniformly over $A$.

The following elementary observation will be used repeatedly, in particular
in the well-posedness theorem below: the Lyapunov inequalities of
Assumption~\ref{ass:main}(A5) are preserved under arbitrary measurable policy
selection, because they are uniform over the action variable.

\begin{definition}[Fixed-policy coefficients and the policy-evaluation residual operator]
\label{def:residual_operator}
Let $\alpha:\mathbb R^d\to A$ be any Borel measurable Markov policy. We
write
\[
\begin{aligned}
b_\alpha(x)&:=b(x,\alpha(x)),\quad
L_\alpha(x)&:=L(x,\alpha(x)),
\end{aligned}
\]
and use the same convention for any other policy symbol. The policy-evaluation
residual operator associated with $\alpha$ is defined by
\[
\mathcal T[v,\alpha]
:=
\lambda v
-
\frac12\operatorname{tr}(\sigma\sigma^\top D_{xx}^2 v)
-
b_\alpha\cdot\nabla_x v
-
L_\alpha .
\]
\end{definition}

\begin{lemma}[Uniform Lyapunov bound under measurable selectors]
\label{lem:lyapunov_selector}
Assume Assumption~\ref{ass:main}. Let
$\alpha:\mathbb R^d\to A$ be any Borel measurable Markov policy. Then
\[
|L_\alpha(x)|
\le
C_L\Psi(x),
\qquad x\in\mathbb R^d,
\]
and
\[
\frac12\operatorname{tr}(\sigma\sigma^\top D_{xx}^2\Psi(x))
+
b_\alpha(x)\cdot\nabla_x\Psi(x)
\le
C_\Psi\Psi(x),
\qquad x\in\mathbb R^d ,
\]
and
\[
\frac12\operatorname{tr}(\sigma\sigma^\top D_{xx}^2\Phi(x))
+
b_\alpha(x)\cdot\nabla_x\Phi(x)
\le
C_\Phi\Phi(x),
\qquad x\in\mathbb R^d .
\]
\end{lemma}
\begin{proof}
Each inequality follows immediately from $\alpha(x)\in A$ and the corresponding
uniform-in-$a$ inequality in Assumption~\ref{ass:main}(A5).
\end{proof}

We now show that the fixed-policy evaluation equation is well posed for
\emph{every} Borel measurable Markov policy. No separate admissibility class
is needed: existence, uniqueness, and the uniform growth bound are proved,
not assumed.

\begin{theorem}[Fixed-policy well-posedness for Borel Markov policies]
\label{thm:wellposed}
Assume Assumption~\ref{ass:main}. Let $\alpha:\mathbb R^d\to A$ be any
Borel measurable Markov policy. Then the fixed-policy equation
\[
\mathcal T[v^\alpha,\alpha]=0
\qquad\text{in }\mathbb R^d
\]
admits a unique strong solution
$v^\alpha\in W^{2,p}_{\mathrm{loc}}(\mathbb R^d)\cap C(\mathbb R^d)$,
for the fixed exponent $p>d$, in the $\Psi$-growth class
$\{v\in C(\mathbb R^d):|v|\le C_v\Psi\text{ for some }C_v>0\}$.
Moreover,
\[
|v^\alpha(x)|
\le
\frac{C_L}{\lambda-C_\Psi}\Psi(x),
\qquad x\in\mathbb R^d,
\]
and this estimate is uniform over all Borel measurable policies
$\alpha:\mathbb R^d\to A$.
\end{theorem}

\begin{proof}
Set $c_\Psi:=C_L/(\lambda-C_\Psi)$ and write
$\mathcal L^\alpha v
:=\frac12\operatorname{tr}(\sigma\sigma^\top D_{xx}^2v)
+b_\alpha\cdot\nabla_x v$, so that
$\mathcal T[v,\alpha]=\lambda v-\mathcal L^\alpha v-L_\alpha$.

\emph{Step 1 (approximating problems).} For $R>0$ consider the Dirichlet
problem
\[
\lambda v_R-\mathcal L^\alpha v_R=L_\alpha
\quad\text{in }B_R,
\qquad
v_R=0
\quad\text{on }\partial B_R .
\]
The principal coefficients are constant and uniformly elliptic
(Assumption~\ref{ass:main}(A4)), the drift $b_\alpha$ is Borel measurable
and bounded on $B_R$ by Assumption~\ref{ass:main}(A2) and the compactness of
$A$, the zeroth-order coefficient $-\lambda$ is negative, and
$L_\alpha\in L^\infty(B_R)\subset L^p(B_R)$. Hence the standard existence
and uniqueness theory for strong solutions of linear uniformly elliptic
equations \citep{gilbarg2001elliptic} provides a unique
$v_R\in W^{2,p}(B_R)\cap W^{1,p}_0(B_R)$; since $p>d$, we also have
$v_R\in C(\overline{B_R})$.

\emph{Step 2 (uniform barrier).} By Lemma~\ref{lem:lyapunov_selector} and
$\lambda>C_\Psi$,
\[
\lambda(c_\Psi\Psi)-\mathcal L^\alpha(c_\Psi\Psi)
\ge
c_\Psi(\lambda-C_\Psi)\Psi
=
C_L\Psi
\ge
L_\alpha
\qquad\text{a.e.\ in }B_R,
\]
and similarly
$\lambda(-c_\Psi\Psi)-\mathcal L^\alpha(-c_\Psi\Psi)
\le-C_L\Psi\le L_\alpha$. Since $\Psi\ge1$ and $v_R=0$ on
$\partial B_R$, the maximum principle for strong solutions on bounded
domains \citep{gilbarg2001elliptic}, applied to
$v_R-c_\Psi\Psi$ and to $-c_\Psi\Psi-v_R$, gives
\[
-c_\Psi\Psi
\le
v_R
\le
c_\Psi\Psi
\qquad\text{in }B_R,
\]
uniformly in $R$ and in the policy $\alpha$.

\emph{Step 3 (compactness and existence).} Fix a ball $B_k$ and take any
$R>k+1$. By Step 2 the family $\{v_R\}$ is uniformly bounded on $B_{k+1}$,
so the interior $W^{2,p}$ estimate for strong solutions
\citep{gilbarg2001elliptic} yields
\[
\sup_{R>k+1}\|v_R\|_{W^{2,p}(B_k)}<\infty .
\]
Exhausting $\mathbb R^d$ by the balls $B_k$, $k\in\mathbb N$, and passing
to a diagonal subsequence $R_j\to\infty$, we obtain
$v^\alpha\in W^{2,p}_{\mathrm{loc}}(\mathbb R^d)$ such that
$v_{R_j}\rightharpoonup v^\alpha$ weakly in $W^{2,p}(B_k)$ and
$v_{R_j}\to v^\alpha$ in $C^1(\overline{B_k})$ for every $k$, using the
compact embedding $W^{2,p}(B_k)\hookrightarrow C^1(\overline{B_k})$ for
$p>d$. Since the equation is linear, the limit passes as follows: the
second-order terms converge weakly in $L^p$ on each ball,
$\nabla_x v_{R_j}\to\nabla_x v^\alpha$ uniformly on compact sets, and,
since $b_\alpha\in L^\infty_{\mathrm{loc}}$,
$b_\alpha\cdot\nabla_x v_{R_j}\to b_\alpha\cdot\nabla_x v^\alpha$
strongly in $L^p_{\mathrm{loc}}$. Hence the limit satisfies
$\lambda v^\alpha-\mathcal L^\alpha v^\alpha=L_\alpha$ almost
everywhere in $\mathbb R^d$. The barrier bound of Step 2 passes to the limit
and gives $|v^\alpha|\le c_\Psi\Psi$ on $\mathbb R^d$; in particular,
$v^\alpha$ belongs to the $\Psi$-growth class, and
$v^\alpha\in C(\mathbb R^d)$ by the Sobolev embedding.

\emph{Step 4 (uniqueness).} Let $u$ and $v$ be two strong solutions in the
$\Psi$-growth class and set $w:=u-v$, so that
$\lambda w-\mathcal L^\alpha w=0$ almost everywhere in $\mathbb R^d$.
Since $w=O(\Psi)$ and $\Psi/\Phi\to0$ at infinity, $w/\Phi\to0$ as
$|x|\to\infty$. For $\varepsilon>0$ put
$w_\varepsilon:=w-\varepsilon\Phi$. By
Lemma~\ref{lem:lyapunov_selector} and $\lambda>C_\Phi$,
\[
\lambda w_\varepsilon-\mathcal L^\alpha w_\varepsilon
\le
-\varepsilon(\lambda-C_\Phi)\Phi
<0
\qquad\text{a.e.\ in }\mathbb R^d .
\]
Because $w/\Phi\to0$ and $\Phi\ge1$, there exists
$M_\varepsilon>0$ such that $w_\varepsilon<0$ for
$|x|\ge M_\varepsilon$, in particular on $\partial B_{M_\varepsilon}$.
Applying the maximum principle for strong solutions
\citep{gilbarg2001elliptic} to $w_\varepsilon$ on
$B_{M_\varepsilon}$ gives $w_\varepsilon\le0$ in $B_{M_\varepsilon}$,
hence $w\le\varepsilon\Phi$ on $\mathbb R^d$. Letting
$\varepsilon\downarrow0$ yields $u\le v$, and exchanging the roles of $u$
and $v$ gives $u=v$.
\end{proof}

In view of Theorem~\ref{thm:wellposed}, every Borel measurable Markov policy
admits a unique exact PDE evaluation in the $\Psi$-growth class
(Remark~\ref{rem:pde_scope}), with a growth
bound that is uniform over the policy. Throughout the theoretical analysis,
the exact policy-evaluation functions $v_n$ and $\hat v_n$ are the unique
solutions provided by Theorem~\ref{thm:wellposed}. The measurability of the
greedy selectors follows from the measurable maximum theorem whenever the
corresponding value gradients are Borel measurable; for neural policies, the
required regularity of the surrogates will be stated separately in
Assumption~\ref{ass:nn_regularity}.

\section{Howard's Policy Improvement Algorithm}

\begin{algorithm}[H]
\caption{Policy Improvement}\label{alg:howard}
\begin{algorithmic}[1]
  \STATE \textbf{Input:} initial Markov policy $a_0(\cdot)$.
  \FOR{$n=0,1,2,\ldots$}
    \STATE \emph{Policy evaluation:} solve for $v_n$ on $\mathbb R^d$
    \[
      \lambda v_n
      -
      \frac12\operatorname{tr}(\sigma\sigma^\top D_{xx}^2v_n)
      -
      b(\cdot,a_n)\cdot\nabla_x v_n
      =
      L(\cdot,a_n).
    \]
    \STATE \emph{Policy improvement:} choose a Borel measurable selector
    \[
      a_{n+1}(x)
      \in
      \mathcal A^\ast\big(x,\nabla_x v_n(x)\big).
    \]
  \ENDFOR
\end{algorithmic}
\end{algorithm}

Howard's policy iteration alternates between evaluating the value of a fixed
policy and improving the policy by a greedy update with respect to the current
value function. Unlike value iteration, which updates the value function
directly through a nonlinear Bellman operator, policy iteration solves a linear
PDE at each evaluation step. This decoupling is useful both analytically and
numerically.

A classical feature of Howard's policy iteration is its monotone improvement
property. In the present PDE framework it takes the following form.

\begin{proposition}[Monotone improvement of the exact iteration]
\label{prop:monotone}
Assume Assumption~\ref{ass:main}, and let $\{(v_n,a_n)\}_{n\ge0}$ be the
exact Howard iteration of Algorithm~\ref{alg:howard}, where each $v_n$ is the
solution provided by Theorem~\ref{thm:wellposed} and $a_{n+1}$ is a Borel
measurable selector of $\mathcal A^\ast(\cdot,\nabla_x v_n(\cdot))$.
Then $v_n\le v_{n+1}$ in $\mathbb R^d$ for every $n\ge0$.
\end{proposition}

\begin{proof}
By the greedy property of the selector,
$L_{a_{n+1}}+b_{a_{n+1}}\cdot\nabla_x v_n
\ge L_{a_n}+b_{a_n}\cdot\nabla_x v_n$
almost everywhere, and hence
\[
\mathcal T[v_n,a_{n+1}]
=
\mathcal T[v_n,a_n]
+
\big(L_{a_n}+b_{a_n}\cdot\nabla_x v_n\big)
-
\big(L_{a_{n+1}}+b_{a_{n+1}}\cdot\nabla_x v_n\big)
\le
0 ,
\]
so $v_n$ is a strong subsolution of the $a_{n+1}$-equation, while
$\mathcal T[v_{n+1},a_{n+1}]=0$. The difference $w:=v_n-v_{n+1}$ therefore
satisfies $\lambda w-\mathcal L^{a_{n+1}}w\le0$ almost everywhere and
$w=O(\Psi)$. Repeating the $\varepsilon\Phi$-perturbation argument of
Step~4 in the proof of Theorem~\ref{thm:wellposed} yields $w\le0$.
\end{proof}

The policy-evaluation step
is linear, while the policy-improvement step is pointwise. In many structured
control problems, especially when the objective is concave in the control
variable, the pointwise maximization is computationally inexpensive.

The main computational bottleneck of Howard's method is the repeated solution of
high-dimensional linear PDEs. Traditional grid-based methods scale poorly with
dimension, which motivates replacing the policy-evaluation PDE solver with a
mesh-free neural PDE solver.

Rather than restating a convergence theorem under hypotheses that would take
us beyond the scope of this paper, we record the convergence behavior of ideal
Howard policy iteration as an explicit assumption, to be invoked in the error
analysis of Section~\ref{sec:closed}.

\begin{assumption}[Exponential convergence of ideal Howard policy iteration]
\label{ass:ideal_conv}
Consider the ideal Howard policy iteration defined in
Algorithm~\ref{alg:howard}. We assume that the sequence of value functions
$\{v_n\}_{n\ge0}$ converges exponentially fast to the optimal value function
$V$: for any compact
set $K\subset\mathbb R^d$, there exist constants $C_K>0$ and
$\rho_{\mathrm{PI}}\in(0,1)$ such that
\[
\|v_n-V\|_{L^\infty(K)}
\le
C_K\rho_{\mathrm{PI}}^n,
\qquad
n\ge0 .
\]
Here $V$ denotes the distinguished solution of the HJB
equation~\eqref{eq:v} to which the ideal PDE iteration is assumed to
converge (cf.\ Remark~\ref{rem:pde_scope}).
\end{assumption}

Exponential (or superlinear) convergence of Howard policy iteration for
continuous-time stochastic control problems has been established in specific
settings; see \citet{Puterman1981,KerimkulovSiskaSzpruch2020} and references
therein for precise statements under the hypotheses of those works. We do not
claim that well-posedness of the policy-evaluation equations and uniqueness of
the optimal value alone imply exponential convergence; instead,
Assumption~\ref{ass:ideal_conv} isolates the convergence property of the ideal
iteration used later in Theorem~\ref{thm:closed}.

\section{Physics-Informed Policy Iteration (PINN-PI)}
\label{sec:pi_hp}

We now introduce a PINN-based variant of Howard policy iteration. At each
policy-iteration step, the value function is approximated by a neural network
trained to minimize the residual of the linear policy-evaluation PDE at sampled
collocation points.

Let $D_{\mathrm{train}}\subset\mathbb R^d$ be a bounded computational domain and
let $\{x_i\}_{i=1}^N\subset D_{\mathrm{train}}$ be collocation points. Although
the theoretical control problem is posed on $\mathbb R^d$, the PINN
implementation is performed on a bounded computational domain. We do not impose
explicit boundary conditions; instead, our analysis and experiments focus on
interior regions of the computational domain.

Given a fixed policy $\tilde a_n(x)$, we approximate the corresponding value
function by a neural network $\tilde v_n(x;w)$ with parameters $w$ and define
the residual loss
\begin{equation}\label{eq:residual}
\begin{aligned}
\mathcal L(w)
:=
\frac1N\sum_{i=1}^N
\Big|
&\lambda \tilde v_n(x_i;w)
-
\frac12
\operatorname{tr}\!\left(
\sigma\sigma^\top D_{xx}^2\tilde v_n(x_i;w)
\right)  \\
&-
b(x_i,\tilde a_n(x_i))\cdot\nabla_x \tilde v_n(x_i;w)
-
L(x_i,\tilde a_n(x_i))
\Big|^2 .
\end{aligned}
\end{equation}

For affine-in-control systems, the policy update often has a closed-form
solution. For more general nonlinear control dependence, we use a lightweight
projected gradient solver for the pointwise maximization problem. We
emphasize that the error analysis of Section~\ref{sec:closed} assumes that
the pointwise maximization in the policy-improvement step is carried out
exactly; optimization errors in this step are not accounted for. Under
Assumptions~\ref{ass:nn_regularity} and~\ref{ass:concavity_dominance}, the
exact greedy maximizer is single-valued on the training domain, so the
following comparison is meaningful: if the implemented update satisfies
$|\tilde a_{n+1}^{\mathrm{num}}(x)-a^\ast(x,\nabla_x\tilde v_n(x))|
\le\eta_n$ uniformly on the training domain, the pointwise policy-error
bound acquires the additional term $\eta_n$, and the final estimate of
Theorem~\ref{thm:closed} an additional error floor proportional to
$\sup_{n\ge0}\eta_n$; we do not pursue this refinement here.

\begin{algorithm}[H]
\caption{Physics-Informed Neural Network Policy Iteration (PINN-PI)}
\label{algo:main}
\begin{algorithmic}[1]
  \STATE \textbf{Input:} initial policy $\tilde a_0(\cdot)$, number of
  collocation points $N$, computational domain $D_{\mathrm{train}}$, initial
  network parameters $w_0$.
  \FOR{$n=0,1,2,\ldots$}
    \STATE \textbf{Collocation sampling:}
    sample $\{x_i\}_{i=1}^N\subset D_{\mathrm{train}}$.
    \STATE \textbf{Policy evaluation:}
    train a neural network $\tilde v_n(x;w_n)$ by minimizing the residual
    loss in~\eqref{eq:residual} with the frozen policy $\tilde a_n$.
    \STATE \textbf{Policy improvement:} choose a Borel measurable selector
    \[
      \tilde a_{n+1}(x)
      \in
      \mathcal A^\ast\big(x,\nabla_x\tilde v_n(x;w_n)\big).
    \]
    \STATE \textbf{If} a stopping criterion is met, stop.
  \ENDFOR
\end{algorithmic}
\end{algorithm}

Recall from Definition~\ref{def:residual_operator} the residual operator
$\mathcal T[\cdot,a]$ associated with a fixed Markov policy
$a:\mathbb R^d\to A$; the exact fixed-policy evaluation under $a$ is the
solution of $\mathcal T[v,a]=0$.

Our analysis compares three families of value functions. The \emph{ideal}
policy-iteration sequence $\{(v_n,a_n)\}_{n\ge0}$ performs exact policy
evaluation followed by greedy improvement,
\begin{equation}\label{eq:vn_def}
\mathcal T[v_n,a_n]=0,
\qquad
a_{n+1}(x)\in\mathcal A^\ast(x,\nabla_x v_n(x)).
\end{equation}
The \emph{PINN} sequence $\{(\tilde v_n,\tilde a_n)\}_{n\ge0}$ replaces each
exact evaluation by a neural surrogate $\tilde v_n$, whose PDE defect is the
residual
\begin{equation}\label{eq:rn_def}
r_n:=\mathcal T[\tilde v_n,\tilde a_n],
\qquad
\tilde a_{n+1}(x)\in\mathcal A^\ast(x,\nabla_x\tilde v_n(x)).
\end{equation}
Here the neural surrogate $\tilde v_n$ is regarded as a globally defined
smooth function on $\mathbb R^d$: a neural network with smooth activations is
defined on the whole space, and training merely selects its parameters. The
greedy update in~\eqref{eq:rn_def} therefore defines a policy on all of
$\mathbb R^d$, where, at points at which the maximizer is not unique, a Borel
measurable selector is chosen (Remark~\ref{rem:measurable} below). Every
policy obtained in this way is Borel measurable, so, by
Theorem~\ref{thm:wellposed}, it admits a unique exact PDE evaluation in the
$\Psi$-growth class; no separate admissibility condition is required. The
error analysis depends on the behavior of $\tilde a_n$ outside the
computational domain only through the uniform bounds of
Theorem~\ref{thm:wellposed} and Lemma~\ref{lem:lyapunov_selector}, which
hold uniformly over all Borel measurable policies, so the estimates below are
uniform with respect to the choice of selector.

Finally, $\hat v_n$ denotes the exact fixed-policy evaluation under the frozen
neural policy $\tilde a_n$,
\begin{equation}\label{eq:hvn_def}
\mathcal T[\hat v_n,\tilde a_n]=0 .
\end{equation}

Throughout the error analysis we write
\[
\delta_n:=\tilde v_n-\hat v_n,
\qquad
\varepsilon_n:=\hat v_n-v_n,
\qquad\text{so that}\qquad
\tilde v_n-v_n=\delta_n+\varepsilon_n .
\]
Here $\delta_n$ is the PINN evaluation error under the frozen neural policy
$\tilde a_n$, and $\varepsilon_n$ is the policy-mismatch error between the
frozen neural policy $\tilde a_n$ and the ideal policy $a_n$.
Section~\ref{sec:closed} bounds both components on interior domains and closes
the resulting error recursion.

\section{Interior error analysis and a closed finite-step error estimate}
\label{sec:closed}

Two structural difficulties drive the analysis. First, no boundary condition is
imposed on $D_{\mathrm{train}}$: the homogeneous fixed-policy equation admits
nonzero solutions on any bounded domain, so the PDE residual alone cannot
control the surrogate (Remark~\ref{rem:necessity}), and some substitute for the
missing boundary information is required. Second, value-gradient errors feed
back into the next policy through the greedy update, so evaluation errors
propagate across policy-iteration steps. We address the first difficulty with
an exponential interior attenuation estimate for the coercive operator
$\lambda-\frac12\operatorname{tr}(\sigma\sigma^\top
D^2_{xx})-b\cdot\nabla_x$ (Lemma~\ref{lem:attenuation}): boundary information
influences the interior only up to a factor decaying exponentially in the
buffer width, so a sufficiently wide training buffer, together with the a
priori amplitude bounds of Assumption~\ref{ass:nn_regularity}, acts as a
substitute for boundary data. We address the second difficulty with a pointwise
gradient-to-policy stability estimate (Proposition~\ref{prop:policy_err}) and
close the recursion on nested domains (Theorem~\ref{thm:closed}).

Throughout this section we fix bounded Lipschitz ambient domains
$\mathcal D\Subset\mathcal D^{+}\subset\mathbb R^d$, with $\mathcal D$
containing all computational domains below and $\mathcal D^{+}$ serving as
the outer domain for the ambient interior estimates; for
$U\subset\mathbb R^d$ and $r>0$ we write
$U\oplus B_r:=\{x:\operatorname{dist}(x,U)<r\}$, with the convention
$U\oplus B_0:=U$, and we set
\[
B_{\mathcal D}:=\Big\|\sup_{a\in A}|b(\cdot,a)|\Big\|_{L^\infty(\mathcal D^{+})}<\infty .
\]
For the reader's convenience, Table~\ref{tab:notation} summarizes the roles
of the main quantities appearing in the error analysis.

\begin{table}[htbp]
\centering
\small
\begin{tabular}{@{}l p{0.78\textwidth}@{}}
\toprule
Symbol & Role \\
\midrule
$\Psi$ & Lyapunov weight controlling the growth of the data and of the exact evaluations \\
$\Phi$ & faster comparison weight, $\Psi/\Phi\to0$; weight of the comparison class \\
$p>d$ & fixed Sobolev exponent for strong solutions and interior estimates \\
$B_{\mathcal D}$ & uniform drift bound on the ambient domain \\
$\kappa_d$ & exponential attenuation rate (Lemma~\ref{lem:attenuation}) \\
$Q_{\mathcal D}$ & a priori bound for the policy-mismatch error $\hat v_n-v_n$ on $\mathcal D$ \\
$M_\delta$ & a priori amplitude bound for the PINN evaluation error $\tilde v_n-\hat v_n$ \\
$G_{\mathcal D}$, $\widetilde G_{\mathcal D}$ & gradient bounds for exact evaluations (constants via $\mathcal D\Subset\mathcal D^{+}$) and neural surrogates \\
$\theta_{\mathcal D}$ & Lipschitz constant of the greedy policy map (Proposition~\ref{prop:policy_err}) \\
$\tau_n$, $\bar\tau$ & $L^p$ residual tolerances (Assumption~\ref{ass:lp_residual}) \\
$\gamma$ & policy-feedback contraction constant (Proposition~\ref{prop:closed_recursion}) \\
\bottomrule
\end{tabular}
\caption{Main quantities in the interior error analysis of
Section~\ref{sec:closed}.}
\label{tab:notation}
\end{table}

\subsection{Stability of the greedy policy map}

\begin{proposition}[Gradient-to-policy stability]
\label{prop:policy_err}
Assume \textup{(A1)--(A3)} and let $M>0$ satisfy
\[
\mu_a>L_{ab}M .
\]
Fix $x\in\mathbb R^d$. Then, for every $z$ with $|z|\le M$, the set
$\mathcal A^\ast(x,z)$ of~\eqref{eq:greedy_map} is a singleton, whose
unique element we denote by $a^\ast(x,z)$, and $a^\ast(x,\cdot)$ is
Lipschitz continuous on this range. More precisely,
\[
|a^\ast(x,z)-a^\ast(x,z')|
\le
\theta |z-z'|,
\qquad |z|,|z'|\le M,
\]
where
\[
\theta
=
\frac{B_a}{\mu_a-L_{ab}M}.
\]
\end{proposition}
\begin{proof}
We first record single-valuedness. For $F_z(a):=L(x,a)+b(x,a)\cdot z$,
Assumptions (A2)--(A3) give, for all $a,a'\in A$,
\[
\left\langle \nabla_aF_z(a)-\nabla_aF_z(a'),\,a-a'\right\rangle
\le
-\big(\mu_a-L_{ab}|z|\big)|a-a'|^2 ,
\]
so for $|z|\le M$ with $\mu_a>L_{ab}M$ the objective $F_z$ is strongly
concave on the convex set $A$ and its maximizer is unique; the notation
$a^\ast(x,z)$ below therefore refers to this unique maximizer.

Let $z,z' \in \mathbb{R}^d$ and denote
\[
a := a^{\ast}(x,z), \qquad a' := a^{\ast}(x,z').
\]
We may assume $a\ne a'$, since the claimed bound is trivial when $a=a'$.
The first-order optimality conditions (variational inequalities) yield
\begin{align}
\left\langle \nabla_a L(x,a) + \partial_a b(x,a)^\top z,  a' - a \right\rangle &\le 0, \label{vi1} \\
\left\langle \nabla_a L(x,a') + \partial_a b(x,a')^\top z',  a - a' \right\rangle &\le 0. \label{vi2}
\end{align}
Adding \eqref{vi1} and \eqref{vi2} gives
\begin{align*}
&\left\langle \nabla_a L(x,a) - \nabla_a L(x,a'),  a' - a \right\rangle  + \left\langle \left[ \partial_a b(x,a) - \partial_a b(x,a') \right]^\top z,  a' - a \right\rangle\\
&\quad+ \left\langle \partial_a b(x,a')^\top (z - z'),  a' - a \right\rangle \le 0.
\end{align*}

By the $\mu_a$-strong concavity of $L(x,\cdot)$, we have
\[
\left\langle \nabla_a L(x,a) - \nabla_a L(x,a'),  a - a' \right\rangle
\le -\mu_a |a - a'|^2.
\]
Substituting this inequality into the previous display and taking absolute values,
we obtain
\begin{align*}
\mu_a |a - a'|^2
&\le \left| \left\langle [ \partial_a b(x,a) - \partial_a b(x,a') ]^\top z,  a - a' \right\rangle \right| \\
&\quad + \left| \left\langle \partial_a b(x,a')^\top (z - z'),  a - a' \right\rangle \right|.
\end{align*}

\paragraph{Estimate of the first term}
By Assumption~\ref{ass:main}(A2), $\partial_a b(x,a)$ is Lipschitz in $a$
uniformly in $x$. Hence,
\begin{align*}
\left| \left\langle [ \partial_a b(x,a) - \partial_a b(x,a') ]^\top z,  a - a' \right\rangle \right|
&\le |\partial_a b(x,a)-\partial_a b(x,a')|\,|z|\,|a-a'| \\
&\le L_{ab}|z|\,|a-a'|^2.
\end{align*}

\paragraph{Estimate of the second term}
Again by Assumption~\ref{ass:main}(A2), $\partial_a b$ is uniformly bounded.
Therefore,
\[
\left| \left\langle \partial_a b(x,a')^\top (z - z'),  a - a' \right\rangle \right|
\le \|\partial_a b(x,a')\|\,|z-z'|\,|a-a'|
\le B_a |z - z'|\,|a-a'|.
\]

Combining the two estimates yields
\[
\mu_a |a - a'|^2
\le L_{ab}|z|\,|a-a'|^2 + B_a |z - z'|\,|a-a'|.
\]
Rearranging,
\[
(\mu_a - L_{ab}|z|)\,|a-a'|^2 \le B_a |z-z'|\,|a-a'|.
\]
Since $|z|,|z'|\le M$ and $\mu_a > L_{ab} M$, we have $\mu_a - L_{ab}|z|\ge \mu_a - L_{ab}M>0$,
and dividing by $|a-a'|$ gives
\[
|a-a'|
\le \frac{B_a}{\mu_a - L_{ab}M}\,|z-z'|
=: \theta\,|z-z'|.
\]
This proves the desired Lipschitz continuity of $a^{\ast}(x,\cdot)$ on bounded sets.
\end{proof}

The Lipschitz continuity of the policy-improvement map is the key structural
property that converts value-gradient errors into policy errors. It is a
pointwise estimate; this will allow us to localize the error analysis to
nested computational domains below. We emphasize that, when $b$ is nonlinear
in the control, single-valuedness of $a^\ast(x,\cdot)$ on the range
$\{|z|\le M\}$ follows from the dominance condition $\mu_a>L_{ab}M$, not
from strong concavity of $L(x,\cdot)$ alone.

\subsection{Uniform bounds for exact policy evaluations}

For the policy-improvement step, control of the value itself is not
sufficient: the greedy update depends on the value gradient $\nabla_x v$. We
therefore upgrade the uniform growth bound of Theorem~\ref{thm:wellposed} to
a local gradient bound for the exact policy-evaluation solutions; the neural
surrogates are not exact PDE solutions, so their gradient control is handled
separately through a network-level regularity condition
(Assumption~\ref{ass:nn_regularity} below).
This step uses standard interior regularity for fixed-policy linear elliptic
equations. We emphasize that no smoothness of the policy $\alpha$ in the state
variable is required: the frozen coefficients $b_\alpha$ and $L_\alpha$ are
merely Borel measurable and locally bounded, and they enter the interior
$W^{2,p}$ estimates invoked below only through their
$L^\infty$ bounds, which are uniform over the compact action set $A$.

\begin{lemma}[Uniform local gradient bound for exact policy evaluations]
\label{lem:exact_local_grad}
Let $D\Subset D'\Subset\mathbb R^d$ be bounded Lipschitz domains.
Assume Assumption~\ref{ass:main}. Let $\{v_n\}_{n\ge0}$ and $\{\hat v_n\}_{n\ge0}$ be the exact
policy-evaluation sequences of~\eqref{eq:vn_def} and~\eqref{eq:hvn_def},
which are well defined for the Borel measurable policies $a_n$ and
$\tilde a_n$ by Theorem~\ref{thm:wellposed}. Then, for the fixed exponent
$p>d$, there exists a constant $C_{D,D',p}>0$, independent of $n$, such that
\[
\sup_{n\ge0}
\left(
\|v_n\|_{W^{2,p}(D)}
+
\|\hat v_n\|_{W^{2,p}(D)}
\right)
\le
C_{D,D',p}.
\]
In particular, there exists a constant $G_D>0$ such that
\[
\sup_{n\ge0}
\left(
\|\nabla_x v_n\|_{L^\infty(D)}
+
\|\nabla_x \hat v_n\|_{L^\infty(D)}
\right)
\le
G_D .
\]
\end{lemma}

\begin{proof}
We prove the estimate for a general Borel measurable Markov policy
$\alpha:\mathbb R^d\to A$. Let $v^\alpha$ be the exact policy-evaluation
solution provided by Theorem~\ref{thm:wellposed}, so that
$\mathcal T[v^\alpha,\alpha]=0$ almost everywhere in $\mathbb R^d$.
Since $\alpha(x)\in A$ and $A$ is compact, Assumption~\ref{ass:main}(A2) gives
\[
\|b_\alpha\|_{L^\infty(D')}
\le
\Big\|\sup_{a\in A}|b(\cdot,a)|\Big\|_{L^\infty(D')}
<\infty ,
\]
with a bound independent of $\alpha$. Similarly, Assumption~\ref{ass:main}(A3)
gives
\[
\|L_\alpha\|_{L^\infty(D')}
\le
\Big\|\sup_{a\in A}|L(\cdot,a)|\Big\|_{L^\infty(D')}
<\infty .
\]
By Theorem~\ref{thm:wellposed}, the exact evaluations are uniformly locally
bounded:
\[
\|v^\alpha\|_{L^\infty(D')}
\le
\frac{C_L}{\lambda-C_\Psi}\sup_{D'}\Psi
=:
C_{D'} .
\]
We now invoke the interior $W^{2,p}$ estimate for strong solutions of linear
uniformly elliptic equations; see \citet{gilbarg2001elliptic},
and \citet{caffarelli1996uniqueness,krylov1987nonlinear} for the
corresponding theory with measurable ingredients. The fixed-policy equation
$\mathcal T[v^\alpha,\alpha]=0$ satisfies, on $D'$, precisely the required
hypotheses: (i) the principal part $-\frac12\operatorname{tr}(\sigma\sigma^\top
D^2_{xx}\,\cdot\,)$ has constant, uniformly elliptic coefficients with
ellipticity bounds $\nu,\Lambda$ by Assumption~\ref{ass:main}(A4); (ii) the
drift coefficient $b_\alpha$ is Borel measurable with
$\|b_\alpha\|_{L^\infty(D')}<\infty$ as shown above; (iii) the zeroth-order
coefficient $\lambda>0$ is constant; and (iv) the source term
$L_\alpha\in L^\infty(D')\subset L^p(D')$. Moreover, by
Theorem~\ref{thm:wellposed}, $v^\alpha$ is a strong
$W^{2,p}_{\mathrm{loc}}$ solution, so the estimate applies directly.
Consequently, the interior estimate yields
\[
\|v^\alpha\|_{W^{2,p}(D)}
\le
C
\left(
\|v^\alpha\|_{L^p(D')}
+
\|L_\alpha\|_{L^p(D')}
\right),
\]
where $C$ depends only on
$d$, $p$, $\lambda$, $\nu$, $\Lambda$,
$\|b_\alpha\|_{L^\infty(D')}$, and $\operatorname{dist}(D,\partial D')$, but
not on the particular policy $\alpha$: since
$\|b_\alpha\|_{L^\infty(D')}\le\|\sup_{a\in A}|b(\cdot,a)|\,\|_{L^\infty(D')}$
uniformly over Borel measurable policies, the constant is uniform over all
Borel measurable policies. Using the uniform bounds above and the boundedness of $D'$,
we obtain
\[
\|v^\alpha\|_{W^{2,p}(D)}
\le
C_{D,D',p}.
\]
Applying this estimate to $\alpha=a_n$ gives the bound for $v_n$, and applying
it to $\alpha=\tilde a_n$ gives the bound for $\hat v_n$.

Finally, since $p>d$, the Sobolev embedding
\[
W^{2,p}(D)\hookrightarrow C^{1,\gamma}(\overline D),
\qquad
\gamma=1-\frac{d}{p}>0,
\]
implies
\[
\|\nabla_x v^\alpha\|_{L^\infty(D)}
\le
C\|v^\alpha\|_{W^{2,p}(D)} .
\]
Therefore,
\[
\sup_{n\ge0}
\left(
\|\nabla_x v_n\|_{L^\infty(D)}
+
\|\nabla_x \hat v_n\|_{L^\infty(D)}
\right)
\le
G_D
\]
for some constant $G_D>0$ independent of $n$.
\end{proof}

Two consequences on the ambient domain will be used repeatedly. First, by
Theorem~\ref{thm:wellposed},
\[
Q_{\mathcal D}
:=
\frac{2C_L}{\lambda-C_\Psi}\sup_{\mathcal D}\Psi
\;\ge\;
\sup_{n\ge0}\|\hat v_n-v_n\|_{L^\infty(\mathcal D)} .
\]
Second, applying Lemma~\ref{lem:exact_local_grad} with $\mathcal D$ as the
inner domain and $\mathcal D^{+}$ as the outer domain, we obtain a constant
$G_{\mathcal D}>0$, depending on $\mathcal D$, $\mathcal D^{+}$, $p$, and
the local coefficient bounds on $\mathcal D^{+}$, such that
\[
\sup_{n\ge0}
\left(
\|\nabla_x v_n\|_{L^\infty(\mathcal D)}
+
\|\nabla_x\hat v_n\|_{L^\infty(\mathcal D)}
\right)
\le
G_{\mathcal D} .
\]

\subsection{Neural surrogate regularity and concavity dominance}

\begin{assumption}[Neural surrogate regularity]
\label{ass:nn_regularity}
Each trained neural surrogate
\[
\tilde v_n(\cdot)=\tilde v_n(\cdot;w_n)
\]
is a globally defined $C^2$ neural-network function on $\mathbb R^d$, so
that the pointwise residual $r_n$ in~\eqref{eq:rn_def} is defined in the
classical sense. Moreover, the trained parameters remain in a bounded
parameter set. The following uniform output and gradient bounds are assumed
only on the training domain: there exist constants
$\widetilde G_{\mathcal D}>0$ and $\widetilde K>0$ such that
\[
\sup_{n\ge0}
\|\nabla_x\tilde v_n\|_{L^\infty(D_{\mathrm{train}})}
\le
\widetilde G_{\mathcal D},
\qquad
\sup_{n\ge0}
\|\tilde v_n\|_{L^\infty(D_{\mathrm{train}})}
\le
\widetilde K .
\]
\end{assumption}

\begin{remark}[Why this assumption is only imposed on the neural surrogates]
The exact policy-evaluation solutions are controlled by PDE regularity: the
preceding local boundedness and elliptic regularity estimates provide uniform
local bounds for $\nabla_x v_n$ and $\nabla_x\hat v_n$ on bounded domains.
Assumption~\ref{ass:nn_regularity} concerns the neural
surrogates, which are not exact PDE solutions. The condition is natural for
fixed smooth neural-network architectures with bounded trained parameters
(e.g., networks with $\tanh$ activations, as used in our experiments, are
$C^\infty$, and their outputs and gradients are uniformly bounded on the
compact set $\overline{D_{\mathrm{train}}}$); boundedness of the trained
parameters can be encouraged by weight decay and enforced by explicit
parameter clipping, norm constraints, or spectral normalization. It is
therefore a network-level regularity condition, not a regularity assumption on
the exact control problem.
\end{remark}

\begin{remark}[Measurability of neural greedy policies]
\label{rem:measurable}
The surrogate $\tilde v_n$ is a globally defined smooth network, so the map
$x\mapsto\nabla_x\tilde v_n(x)$ is continuous, hence Borel measurable, on
all of $\mathbb R^d$; the quantitative bounds of
Assumption~\ref{ass:nn_regularity} are needed only on $D_{\mathrm{train}}$.
Since $A$ is compact and the objective
\[
a\mapsto L(x,a)+b(x,a)\cdot\nabla_x\tilde v_n(x)
\]
is continuous in $a$ and Borel measurable in $x$, the measurable maximum
theorem implies that the correspondence
\[
x\longmapsto
\mathcal A^\ast\big(x,\nabla_x\tilde v_n(x)\big)
\]
admits a Borel measurable selector on $\mathbb R^d$; we choose one such
selector and denote it by $\tilde a_{n+1}$. Under the
concavity-dominance condition of
Assumption~\ref{ass:concavity_dominance}, this selector is unique at every
point where the relevant value gradients lie in the range $M_{\mathcal D}$,
in particular on $D_{\mathrm{train}}$ and hence on all nested domains used
in the error analysis; outside this region any measurable selector may be
chosen, and, by Theorem~\ref{thm:wellposed}, the resulting policy still
admits a unique exact PDE evaluation.
\end{remark}

\begin{assumption}[Concavity dominance on the ambient domain]
\label{ass:concavity_dominance}
Let $G_{\mathcal D}$ be the exact-gradient bound above and let
$\widetilde G_{\mathcal D}$ be the neural-gradient bound from
Assumption~\ref{ass:nn_regularity}. Define
\[
M_{\mathcal D}:=\max\{G_{\mathcal D},\widetilde G_{\mathcal D}\}.
\]
We assume that
\[
\mu_a>L_{ab}M_{\mathcal D} .
\]
In particular, for every $x\in\mathcal D$ and every $z$ with
$|z|\le M_{\mathcal D}$, the map
\[
a\mapsto L(x,a)+b(x,a)\cdot z
\]
is strongly concave on $A$.
\end{assumption}

\begin{remark}[Control-affine dynamics]
If $b$ is affine in the control variable, then $L_{ab}=0$. In that common case,
Assumption~\ref{ass:concavity_dominance} reduces to the strong concavity
of $L(x,\cdot)$ and is automatically satisfied whenever $\mu_a>0$.
\end{remark}

Under Assumption~\ref{ass:concavity_dominance}, the pointwise
policy-improvement map is Lipschitz on the gradient range encountered on the
ambient domain. By Proposition~\ref{prop:policy_err}, for
\[
\theta_{\mathcal D}
:=
\frac{B_a}{\mu_a-L_{ab}M_{\mathcal D}},
\]
we have, pointwise,
\begin{equation}\label{eq:pointwise_policy}
|a^\ast(x,\nabla_x\tilde v(x))-a^\ast(x,\nabla_x v(x))|
\le
\theta_{\mathcal D}\,
|\nabla_x\tilde v(x)-\nabla_x v(x)|
\end{equation}
whenever $|\nabla_x\tilde v(x)|,|\nabla_x v(x)|\le M_{\mathcal D}$. This
estimate is the mechanism that converts value-gradient errors into policy
errors.

\subsection{Exponential interior attenuation}

The next lemma quantifies how the influence of the (unprescribed) boundary
data decays into the interior. Coercivity, $\lambda>0$, is essential.

\begin{lemma}[Exponential interior attenuation with an $L^p$ source]
\label{lem:attenuation}
Let $U\subset\mathcal D$ be a bounded domain and let
$w\in C(\overline U)\cap W^{2,p}_{\mathrm{loc}}(U)$ be a strong solution of
\[
\lambda w-\tfrac12\operatorname{tr}(\sigma\sigma^\top D^2_{xx}w)
-b\cdot\nabla_x w=r
\qquad\text{a.e.\ in }U,
\]
where $b$ is Borel measurable with $\|b\|_{L^\infty(U)}\le B_{\mathcal D}$ and
$r\in L^p(U)$, with the fixed exponent $p>d$. Let $\kappa>0$ satisfy
\begin{equation}\label{eq:kappa}
\tfrac12\Lambda\kappa^2+B_{\mathcal D}\kappa\le\lambda,
\qquad\text{e.g.}\quad
\kappa=\frac{-B_{\mathcal D}+\sqrt{B_{\mathcal D}^2+2\Lambda\lambda}}{\Lambda},
\end{equation}
and set $\kappa_d:=\kappa/\sqrt d$. Then there exists a constant
$C_{p,\mathcal D}>0$, depending only on
$d,p,\lambda,\nu,\Lambda,B_{\mathcal D}$ and the fixed ambient domain
$\mathcal D$, such that, for every $x_0\in U$,
\begin{equation}\label{eq:lp_attenuation}
|w(x_0)|
\le
C_{p,\mathcal D}\,\|r\|_{L^p(U)}
+
2d\,e^{-\kappa_d\operatorname{dist}(x_0,\partial U)}\,
\|w\|_{L^\infty(U)} .
\end{equation}
Moreover, whenever $r\in L^\infty(U)$, the sharper source estimate
\begin{equation}\label{eq:linfty_attenuation}
|w(x_0)|
\le
\frac{\|r\|_{L^\infty(U)}}{\lambda}
+
2d\,e^{-\kappa_d\operatorname{dist}(x_0,\partial U)}\,
\|w\|_{L^\infty(U)}
\end{equation}
also holds.
\end{lemma}

\begin{proof}
Choose once and for all a ball $B_R$ with
$\mathcal D\Subset B_R$. Extend $b$ from $U$ to a bounded Borel vector field
$\bar b$ on $B_R$ with $\|\bar b\|_{L^\infty(B_R)}\le B_{\mathcal D}$, and
extend $r$ by zero outside $U$. Let $z$ be the zero-Dirichlet strong solution of
\[
\lambda z-
\tfrac12\operatorname{tr}(\sigma\sigma^\top D^2_{xx}z)
-\bar b\cdot\nabla_xz
=r\,\mathbf 1_U
\quad\text{in }B_R,
\qquad
z=0
\quad\text{on }\partial B_R .
\]
The standard global $W^{2,p}$ estimate on the fixed ball $B_R$ for a constant uniformly elliptic principal part and bounded lower-order coefficients~\citep{gilbarg2001elliptic}, followed by
$W^{2,p}(B_R)\hookrightarrow C^{1,\alpha}(\overline{B_R})$ for $p>d$, gives
\begin{equation}\label{eq:source_corrector}
\|z\|_{L^\infty(B_R)}
\le
C_{p,\mathcal D}\,\|r\|_{L^p(U)} .
\end{equation}
Here and below the constant may increase from line to line but depends only on
the displayed structural and ambient-domain quantities. This is the only step
at which the finite-$p$ source norm enters.

Set $h:=w-z$ on $U$. Since the extended equation for $z$ agrees with that of
$w$ inside $U$, the function $h$ satisfies the homogeneous equation
\[
\lambda h-
\tfrac12\operatorname{tr}(\sigma\sigma^\top D^2_{xx}h)
-b\cdot\nabla_xh=0
\qquad\text{a.e.\ in }U.
\]
Write $\rho_0:=\operatorname{dist}(x_0,\partial U)$ and fix
$0<t<\rho_0$. On $B_t(x_0)\Subset U$ use the same coordinatewise $\cosh$
barrier as in the sup-norm-source argument, now with zero source:
\[
\mathfrak b(x)
:=
\|h\|_{L^\infty(U)}
\sum_{i=1}^d
\frac{\cosh\!\big(\kappa(x_i-x_{0,i})\big)}
{\cosh(\kappa t/\sqrt d)} .
\]
On $\partial B_t(x_0)$ the sum is at least one, and
\eqref{eq:kappa} gives
\[
\lambda\mathfrak b-
\tfrac12\operatorname{tr}(\sigma\sigma^\top D^2_{xx}\mathfrak b)
-b\cdot\nabla_x\mathfrak b\ge0 .
\]
The maximum principle applied to $h-\mathfrak b$ and to $-h-\mathfrak b$
therefore yields
\[
|h(x_0)|
\le
2d\,e^{-\kappa_dt}\|h\|_{L^\infty(U)} .
\]
Letting $t\uparrow\rho_0$ and using
$\|h\|_{L^\infty(U)}\le
\|w\|_{L^\infty(U)}+\|z\|_{L^\infty(B_R)}$, we obtain
\[
|w(x_0)|
\le
\|z\|_{L^\infty(B_R)}
+2d e^{-\kappa_d\rho_0}
\big(\|w\|_{L^\infty(U)}+\|z\|_{L^\infty(B_R)}\big).
\]
Absorbing the last $z$-term into the first one by
\eqref{eq:source_corrector} proves \eqref{eq:lp_attenuation}.

If in addition $r\in L^\infty(U)$, repeat the same $\cosh$-barrier argument
directly for $w$, with the barrier
\[
\frac{\|r\|_{L^\infty(U)}}{\lambda}
+
\|w\|_{L^\infty(U)}
\sum_{i=1}^d
\frac{\cosh\!\big(\kappa(x_i-x_{0,i})\big)}
{\cosh(\kappa t/\sqrt d)} .
\]
The constant term contributes exactly $\|r\|_{L^\infty(U)}$ under the
coercive operator, and \eqref{eq:kappa} controls the remaining terms. Letting
$t\uparrow\operatorname{dist}(x_0,\partial U)$ gives
\eqref{eq:linfty_attenuation}.
\end{proof}

\begin{remark}[Necessity of outer information]
\label{rem:necessity}
Some outer control is necessary: for $r\equiv0$ the homogeneous equation
admits nonzero solutions of exponential type (e.g.\
$w(x)=e^{\sqrt{2\lambda}\,x_1}$ when $\sigma\sigma^\top=I$ and $b\equiv0$) on
any bounded domain, so a residual-only interior bound is impossible without
boundary conditions. Lemma~\ref{lem:attenuation} shows that the influence of
the missing boundary data decays exponentially with distance, which is a
natural form of outer control and, as we show next, it suffices.
\end{remark}

The rate in~\eqref{eq:kappa} is sharp already for $r\equiv0$: for constant
drift in one dimension the slower homogeneous root is exactly $\kappa$,
confirmed to four digits in Section~\ref{sec:exp_buffer}, and the worst case
it encodes is a drift carrying boundary information inward.

Two sources of slack remain. The factor $\sqrt d$ in $\kappa_d$ and the
prefactor $2d$ come from the coordinatewise $\cosh$ barrier rather than from
the equation: in two dimensions the realized rate is $\kappa$, not
$\kappa/\sqrt2$. More consequentially,
Assumption~\ref{ass:main}(A2) enters only through the sup norm
$B_{\mathcal D}$, attained near the corners of the domain in the least
favourable direction; for a stabilizing closed loop the drift points inward
everywhere and the realized attenuation is one to three orders of magnitude
faster than $e^{-\kappa_d\operatorname{dist}}$ permits
(Table~\ref{tab:kappa}), so the buffer widths Theorem~\ref{thm:closed}
requires far exceed those needed in practice. A barrier adapted to the
dissipativity would close much of this gap; we do not pursue it here.

\subsection{PINN evaluation error without boundary conditions}

\begin{assumption}[$L^p$ residual tolerance]
\label{ass:lp_residual}
For the fixed exponent $p>d$, there are constants $\tau_n>0$ with
\[
\bar\tau:=\sup_{n\ge0}\tau_n<\infty
\]
such that
\[
\|r_n\|_{L^p(D_{\mathrm{train}})}\le \tau_n,
\qquad n=0,1,2,\ldots .
\]
No consistency between $\tilde v_n$ and the exact frozen-policy evaluation
$\hat v_n$ is assumed; the only outer information used below is the a priori
amplitude bound of Assumption~\ref{ass:nn_regularity}.
\end{assumption}

\begin{remark}[Empirical versus continuous residuals]
\label{rem:empirical_residual}
In practice, Algorithm~\ref{algo:main} minimizes the empirical squared
collocation loss~\eqref{eq:residual}, whereas
Assumption~\ref{ass:lp_residual} concerns the continuous $L^p$ residual on
$D_{\mathrm{train}}$, with a finite exponent $p>d$. Thus the assumption is
still a generalization condition, but it no longer requires pointwise control
of the residual. The restriction $p>d$ is structural: it permits the interior
embedding $W^{2,p}\hookrightarrow C^{1,\alpha}$ needed to control the greedy
policy update in the sup norm.

For uniform fresh validation points $X_j\sim\mathrm{Unif}(D_{\mathrm{train}})$,
a natural Monte Carlo diagnostic is
\[
\widehat\tau_n
:=
|D_{\mathrm{train}}|^{1/p}
\left(
\frac1{N_{\mathrm{val}}}
\sum_{j=1}^{N_{\mathrm{val}}}|r_n(X_j)|^p
\right)^{1/p}.
\]
Quantifying the finite-sample generalization gap between this estimator and the
continuous $L^p$ norm is outside the scope of the present paper. Unlike a
sup-norm tolerance, however, the theoretical residual quantity is a finite
moment and can be estimated directly from fresh collocation samples.
\end{remark}

By Assumption~\ref{ass:nn_regularity} and Theorem~\ref{thm:wellposed}, the
PINN evaluation error is a priori bounded:
\[
M_\delta
:=
\widetilde K+\frac{C_L}{\lambda-C_\Psi}\sup_{\mathcal D}\Psi
\;\ge\;
\sup_{n\ge0}\|\tilde v_n-\hat v_n\|_{L^\infty(D_{\mathrm{train}})} .
\]

\begin{proposition}[Interior PINN evaluation error from an $L^p$ residual]
\label{prop:delta_closed}
Assume Assumptions~\ref{ass:main}, \ref{ass:nn_regularity},
and~\ref{ass:lp_residual}, with $D_{\mathrm{train}}\subset\mathcal D$. Let
$U\Subset D_{\mathrm{train}}$ with
$\varrho:=\operatorname{dist}(U,\partial D_{\mathrm{train}})\ge2$. Then
\[
\|\delta_n\|_{L^\infty(U)}
\le
C_{p,\mathcal D}\,\tau_n
+
2dM_\delta\,e^{-\kappa_d\varrho},
\]
and
\[
\|\nabla_x\delta_n\|_{L^\infty(U)}
\le
C\big(\tau_n+M_\delta\,e^{-\kappa_d(\varrho-1)}\big),
\]
where $C$ depends only on $d,p,\lambda,\nu,\Lambda,B_{\mathcal D}$ and the
fixed ambient domain $\mathcal D$.
\end{proposition}

\begin{proof}
Since $\tilde v_n\in C^2(\overline{D_{\mathrm{train}}})$ satisfies
$\mathcal T[\tilde v_n,\tilde a_n]=r_n$ pointwise while $\hat v_n$ is a
strong $W^{2,p}_{\mathrm{loc}}$ solution of
$\mathcal T[\hat v_n,\tilde a_n]=0$
(Theorem~\ref{thm:wellposed}), the difference
$\delta_n=\tilde v_n-\hat v_n\in
C(\overline{D_{\mathrm{train}}})\cap
W^{2,p}_{\mathrm{loc}}(D_{\mathrm{train}})$ is a strong solution of
\[
\lambda\delta_n
-
\frac12\operatorname{tr}(\sigma\sigma^\top D_{xx}^2\delta_n)
-
b_{\tilde a_n}\cdot\nabla_x\delta_n
=
r_n
\qquad\text{a.e.\ in }D_{\mathrm{train}},
\]
and $\|\delta_n\|_{L^\infty(D_{\mathrm{train}})}\le M_\delta$. The first
estimate follows from Lemma~\ref{lem:attenuation} with
$U=D_{\mathrm{train}}$ and
$\|r_n\|_{L^p(D_{\mathrm{train}})}\le\tau_n$.

For the gradient bound, fix $x\in U$. Since $\varrho\ge2$,
$B_1(x)\Subset D_{\mathrm{train}}$. The interior $W^{2,p}$ estimate and the
embedding $W^{2,p}(B_{1/2}(x))\hookrightarrow C^{1,\alpha}(B_{1/2}(x))$ give
\[
\|\nabla_x\delta_n\|_{L^\infty(B_{1/2}(x))}
\le
C\Big(
\|\delta_n\|_{L^p(B_1(x))}
+
\|r_n\|_{L^p(B_1(x))}
\Big).
\]
The local residual term is bounded by $\tau_n$. Moreover every point of
$B_1(x)$ lies at distance at least $\varrho-1$ from
$\partial D_{\mathrm{train}}$, so the first estimate yields
\[
\|\delta_n\|_{L^\infty(B_1(x))}
\le
C_{p,\mathcal D}\tau_n
+2dM_\delta e^{-\kappa_d(\varrho-1)}.
\]
Since $|B_1|$ is fixed, this also controls the $L^p(B_1(x))$ norm. Taking the
supremum over $x\in U$ proves the gradient estimate.
\end{proof}

Proposition~\ref{prop:delta_closed} shows that value-level accuracy of the
surrogate on interior domains is \emph{derived} from the residual, the a
priori bounds, and the buffer width; no outer-domain consistency between
$\tilde v_n$ and $\hat v_n$ is assumed.

\subsection{A closed error recursion and the main theorem}

We now iterate the argument across policy-iteration steps. Fix an inner domain
$D$, a layer width $\rho\ge2$, a reserve buffer $\varrho_0\ge2$, and a number
of steps $n$, and define the nested domains
\[
D^{(k)}
:=
D\oplus B_{(n-k)\rho},
\qquad k=0,1,\ldots,n,
\]
so that $D^{(n)}=D$ and, directly from the definition of the neighborhoods,
\[
D^{(k+1)}\oplus B_\rho\subset D^{(k)},
\qquad k=0,1,\ldots,n-1;
\]
this inclusion is the only geometric property of the nested family used
below. We assume the geometry
\begin{equation}\label{eq:geometry}
D\Subset D_{\mathrm{train}}\subset\mathcal D,
\qquad
\operatorname{dist}(D,\partial D_{\mathrm{train}})\ge n\rho+\varrho_0,
\end{equation}
so that $D^{(0)}$ together with a $\varrho_0$-buffer is contained in
$D_{\mathrm{train}}$.

\begin{proposition}[Closed gradient recursion]
\label{prop:closed_recursion}
Assume Assumptions~\ref{ass:main}, \ref{ass:nn_regularity},
\ref{ass:concavity_dominance}, and~\ref{ass:lp_residual}, together with the
geometry~\eqref{eq:geometry}. Define
\[
G_k:=\|\nabla_x\tilde v_k-\nabla_x v_k\|_{L^\infty(D^{(k)})},
\qquad
\gamma:=C_0\big(1+\lambda^{-1}\big)\big(B_aG_{\mathcal D}+L_a\big)\theta_{\mathcal D},
\]
where $C_0$ depends only on $d,p,\lambda,\nu,\Lambda$, and $B_{\mathcal D}$.
Then, for $k=1,\ldots,n$,
\[
G_k
\le
\gamma\,G_{k-1}
+
C\Big(
\bar\tau
+
Q_{\mathcal D}\,e^{-\kappa_d(\rho-1)}
+
M_\delta\,e^{-\kappa_d(\varrho_0-1)}
\Big).
\]
\end{proposition}

\begin{proof}
Fix $k\ge1$ and decompose
$\nabla_x(\tilde v_k-v_k)=\nabla_x\delta_k+\nabla_x\varepsilon_k$.

\emph{Step 1 ($\delta_k$).} Since $D^{(k)}\subset D^{(0)}$ and
$\operatorname{dist}(D^{(0)},\partial D_{\mathrm{train}})\ge\varrho_0$,
Proposition~\ref{prop:delta_closed} gives
\[
\|\nabla_x\delta_k\|_{L^\infty(D^{(k)})}
\le
C\big(\bar\tau+M_\delta\,e^{-\kappa_d(\varrho_0-1)}\big).
\]

\emph{Step 2 (policy error and the $\varepsilon_k$-equation).} On
$D^{(k-1)}\subset D_{\mathrm{train}}$ both $\nabla_x\tilde v_{k-1}$ and
$\nabla_x v_{k-1}$ lie in the gradient range $M_{\mathcal D}$, so the
pointwise stability estimate~\eqref{eq:pointwise_policy} gives
\[
|\tilde a_k(x)-a_k(x)|
\le
\theta_{\mathcal D}\,
|\nabla_x\tilde v_{k-1}(x)-\nabla_x v_{k-1}(x)|
\le
\theta_{\mathcal D}\,G_{k-1},
\qquad x\in D^{(k-1)} .
\]
Subtracting $\mathcal T[\hat v_k,\tilde a_k]=0$ and $\mathcal T[v_k,a_k]=0$
and freezing the drift at $\tilde a_k$ shows that
$\varepsilon_k=\hat v_k-v_k\in W^{2,p}_{\mathrm{loc}}(\mathbb R^d)$ is a
strong solution of
\[
\lambda\varepsilon_k
-
\frac12\operatorname{tr}(\sigma\sigma^\top D_{xx}^2\varepsilon_k)
-
b_{\tilde a_k}\cdot\nabla_x\varepsilon_k
=
R_k
\qquad\text{a.e.\ in }\mathbb R^d,
\]
where
$R_k:=[b_{\tilde a_k}-b_{a_k}]\cdot\nabla_x v_k+[L_{\tilde a_k}-L_{a_k}]$.
By Assumption~\ref{ass:main}(A2)--(A3) and
$\|\nabla_x v_k\|_{L^\infty(\mathcal D)}\le G_{\mathcal D}$,
\[
\|R_k\|_{L^\infty(D^{(k-1)})}
\le
(B_aG_{\mathcal D}+L_a)\,\theta_{\mathcal D}\,G_{k-1}.
\]

\emph{Step 3 ($\varepsilon_k$, value).} For $x_0\in D^{(k)}\oplus B_1$ we have
$B_{\rho-1}(x_0)\subset D^{(k-1)}$, so the sharper $L^\infty$-source estimate~\eqref{eq:linfty_attenuation} on
$B_{\rho-1}(x_0)$, combined with Step 2 and
$\|\varepsilon_k\|_{L^\infty(\mathcal D)}\le Q_{\mathcal D}$, yields
\[
|\varepsilon_k(x_0)|
\le
\frac{(B_aG_{\mathcal D}+L_a)\theta_{\mathcal D}}{\lambda}\,G_{k-1}
+
2d\,Q_{\mathcal D}\,e^{-\kappa_d(\rho-1)} .
\]

\emph{Step 4 ($\varepsilon_k$, gradient).} Applying the unit-ball interior
$W^{2,p}$ estimate as in Proposition~\ref{prop:delta_closed}, with the value
bound of Step 3 on $D^{(k)}\oplus B_1$ and the source bound of Step 2,
\[
\|\nabla_x\varepsilon_k\|_{L^\infty(D^{(k)})}
\le
C\big(1+\lambda^{-1}\big)(B_aG_{\mathcal D}+L_a)\theta_{\mathcal D}\,G_{k-1}
+
C\,Q_{\mathcal D}\,e^{-\kappa_d(\rho-1)} .
\]
Adding Step 1 and Step 4 proves the claim with
$\gamma:=C_0(1+\lambda^{-1})(B_aG_{\mathcal D}+L_a)\theta_{\mathcal D}$.
\end{proof}

\begin{theorem}[Closed finite-step interior error estimate for PINN-PI]
\label{thm:closed}
Assume the hypotheses of Proposition~\ref{prop:closed_recursion},
Assumption~\ref{ass:ideal_conv} applied with $K=\overline D$ (we write
$C_D:=C_{\overline D}$ for the resulting constant), and the contraction
condition
\[
\gamma<1 .
\]
Then, for all $n\ge1$ satisfying the geometry~\eqref{eq:geometry},
\[
\|\tilde v_n-V\|_{L^\infty(D)}
\le
C\Big(
\bar\tau
+
\gamma^{\,n-1}G_0
+
Q_{\mathcal D}\,e^{-\kappa_d(\rho-1)}
+
M_\delta\,e^{-\kappa_d(\varrho_0-1)}
\Big)
+
C_{D}\rho_{\mathrm{PI}}^{\,n},
\]
where $G_0\le G_{\mathcal D}+\widetilde G_{\mathcal D}$ and $C$ depends
only on the structural and ambient-domain constants
$d$, $p$, $\lambda$, $\nu$, $\Lambda$, $B_{\mathcal D}$, $B_a$, $L_a$,
$G_{\mathcal D}$, $\widetilde G_{\mathcal D}$, $\theta_{\mathcal D}$, and
$(1-\gamma)^{-1}$; in particular, $C$ is independent of $n$, $\rho$,
$\varrho_0$, and the residual tolerances. Moreover, writing
$\log_+t:=\max\{\log t,0\}$ and recalling $\bar\tau>0$ from
Assumption~\ref{ass:lp_residual}, the choices
\[
\rho
=
\max\Big\{2,\;1+\kappa_d^{-1}\log_+\!\big(Q_{\mathcal D}/\bar\tau\big)\Big\},
\qquad
\varrho_0
=
\max\Big\{2,\;1+\kappa_d^{-1}\log_+\!\big(M_\delta/\bar\tau\big)\Big\},
\]
yield
\[
\|\tilde v_n-V\|_{L^\infty(D)}
\le
C\big(\bar\tau+\gamma^{\,n-1}+\rho_{\mathrm{PI}}^{\,n}\big)
\qquad\text{whenever}\quad
\operatorname{dist}(D,\partial D_{\mathrm{train}})
\ge
n\rho+\varrho_0 .
\]
\end{theorem}

\begin{proof}
Iterating Proposition~\ref{prop:closed_recursion} and using $\gamma<1$,
\[
G_{n-1}
\le
\gamma^{\,n-1}G_0
+
\frac{C}{1-\gamma}
\Big(\bar\tau+Q_{\mathcal D}\,e^{-\kappa_d(\rho-1)}
+M_\delta\,e^{-\kappa_d(\varrho_0-1)}\Big).
\]
Decompose $\tilde v_n-V=\delta_n+\varepsilon_n+(v_n-V)$ on $D$. The term
$\delta_n$ is bounded by Proposition~\ref{prop:delta_closed}; the term
$\varepsilon_n$ is bounded by Step 3 of the proof of
Proposition~\ref{prop:closed_recursion}, applied with $k=n$ and
$x_0\in D\subset D^{(n)}\oplus B_1$, together with the bound on $G_{n-1}$; the
last term is bounded by Assumption~\ref{ass:ideal_conv}. Collecting constants, and
noting that summing the geometric series contributes the factor
$(1-\gamma)^{-1}$ to $C$, gives the first display. For the second display,
consider each exponential term separately: if $Q_{\mathcal D}\le \bar\tau$,
then $Q_{\mathcal D}e^{-\kappa_d(\rho-1)}\le \bar\tau$ trivially, while
otherwise the choice of $\rho$ gives
$\rho-1\ge\kappa_d^{-1}\log(Q_{\mathcal D}/\bar\tau)$ and again
$Q_{\mathcal D}e^{-\kappa_d(\rho-1)}\le \bar\tau$; the term involving
$M_\delta$ is handled identically. Since
$G_0\le G_{\mathcal D}+\widetilde G_{\mathcal D}$ is an ambient constant,
absorbing it into $C$ gives the second display.
\end{proof}

Theorem~\ref{thm:closed} is a closed error estimate: no
consistency between the surrogate and the exact frozen-policy evaluation is
assumed, and the policy-mismatch error is absorbed by the recursion. The
price is threefold. First, the estimate is not residual-only: besides the
finite-$L^p$ tolerance (Remark~\ref{rem:empirical_residual}) it uses the a
priori amplitude bound $M_\delta$ of Assumption~\ref{ass:nn_regularity}.
Second, the feedback constant $\gamma$ must be smaller than one; being
proportional to $\theta_{\mathcal D}=B_a/(\mu_a-L_{ab}M_{\mathcal D})$, this
holds for sufficiently strong control concavity $\mu_a$, in analogy with
contraction conditions in approximate policy iteration
\citep{KerimkulovSiskaSzpruch2020}. We do not claim that a large discount
alone ensures $\gamma<1$: the constants $C_0$ and $G_{\mathcal D}$ may
themselves depend on $\lambda$, and $C_0$, originating from interior
elliptic estimates, is not explicit in general.

Third, the buffer must grow linearly in the number of policy-iteration steps
and logarithmically in the target tolerance. This quantifies the empirical
practice of training on a domain larger than the evaluation region, but on a
fixed domain it also caps the admissible number of steps at
$n\le(\operatorname{dist}(D,\partial D_{\mathrm{train}})-\varrho_0)/\rho$:
the estimate is \emph{finite-step}, quantifying the accuracy
$C(\bar\tau+\gamma^{\,n^*-1}+\rho_{\mathrm{PI}}^{\,n^*})$ reachable within
that budget, with
$n^*\asymp\operatorname{dist}(D,\partial D_{\mathrm{train}})/\rho$, and it
does not yield convergence as $n\to\infty$. An asymptotic statement would
require training domains growing with $n$ together with uniform control of
the domain-dependent constants, which lies beyond the present scope.
Finally, the generated policies need only be Borel measurable, their exact
PDE evaluations being provided by Theorem~\ref{thm:wellposed}
(Remark~\ref{rem:pde_scope}), while convergence of the ideal iteration
remains an assumption (Assumption~\ref{ass:ideal_conv}): the theorem closes the
neural error recursion, not the ideal policy-iteration theory.

\section{Experiments}\label{sec:experiments}

The experiments first test the measurable components of the interior error
analysis on an exact-reference LQR problem: boundary attenuation
(Lemma~\ref{lem:attenuation}), the residual-controlled error floor
(Proposition~\ref{prop:delta_closed}), and the feedback recursion
(Proposition~\ref{prop:closed_recursion}). We then compare fixed-policy and
direct nonlinear HJB training, study dimensional scaling, and examine the
resulting feedback on nonlinear and high-dimensional benchmarks
(Sections~\ref{sec:exp_nonlinear}--\ref{sec:exp_posterior}), which leave the
exact-reference setting.

\subsection{Setup, reference solutions, and measured quantities}
\label{sec:exp_setup}

\paragraph{Benchmark: bounded-control stochastic LQR}
\begin{gather*}
  \de X_t=(AX_t+Bu_t)\,\de t+\sigma\,\de W_t,
  \qquad
  L(x,u)=-x^\top Qx-u^\top Ru,
  \\
  u_t\in[-u_{\max},u_{\max}]^m ,
\end{gather*}
with $m=d$, $Q=I_d$, $R=I_m$, $\lambda=0.5$, and $\sigma=\sigma_0I_d$,
$\sigma_0=0.25$, so $\nu=\Lambda=0.0625$; Section~\ref{sec:exp_buffer} also
varies $\lambda\in\{0.5,2\}$ and $\sigma_0\in\{0.25,0.5,1\}$.
$A:=d^{-1/2}Z-I_d$ with $Z$ having i.i.d.\ Gaussian entries, resampled until
Hurwitz; $B$ likewise, which keeps the drift $O(1)$ in $d$. The evaluation and training
domains are the cubes
\[
D=[-L_{\mathrm{eval}},L_{\mathrm{eval}}]^d,
\quad
D_{\mathrm{train}}=[-L_{\mathrm{train}},L_{\mathrm{train}}]^d,
\quad
L_{\mathrm{eval}}=1,\;\; L_{\mathrm{train}}=2,
\]
unless stated otherwise; shrinking or widening the training box always means
varying $L_{\mathrm{train}}$ with the evaluation domain $D$ held fixed. The
physical benchmarks of Section~\ref{sec:exp_nonlinear} scale these cubes per
coordinate (\ref{app:benchmarks}).

The control set is compact, as Assumption~\ref{ass:main} requires, with
\begin{gather*}
  u_{\max}
  :=
  1.25\cdot\sup_{\|x\|_\infty\le L_{\mathrm{train}}}\|u^\star(x)\|_\infty
  =
  1.25\,L_{\mathrm{train}}\max_i\sum_j|K^\star_{ij}| ,
  \\
  K^\star=R^{-1}B^\top P,
\end{gather*}
so the optimal feedback uses at most $0.8$ of the bound on
$D_{\mathrm{train}}$, and $0.4$ on the smaller domain $D$; saturation begins
only at $|x|_\infty\approx1.25\,L_{\mathrm{train}}$, outside
$D_{\mathrm{train}}$. The constrained and unconstrained
Hamiltonians therefore coincide on $D_{\mathrm{train}}$, and the
unconstrained $V^\star$ solves the \emph{constrained} HJB equation there
exactly (measured residual $1.7\times10^{-7}$, float32 round-off). It is not
the value function of the constrained problem globally, since trajectories
may reach the saturation region; but the two agree on $D$ to $10^{-4}$
relative, as Section~\ref{sec:exp_buffer} verifies against a grid solve of
the constrained equation. This is the only design in which the
theory's assumptions and an exact reference coexist: an active control
constraint on the training domain would invalidate the closed-form solution.

\paragraph{Reference solutions}
The exact value is $V^\star(x)=-x^\top Px-c$ with optimal feedback
$u^\star(x)=-R^{-1}B^\top Px$, where $P$ solves
the discounted Riccati equation $A^\top P+PA-PBR^{-1}B^\top P-\lambda P+Q=0$
and $c=\lambda^{-1}\tr(\sigma\sigma^\top P)$; all errors are computed against
$V^\star$ \emph{including} $c$. For a linear policy $u=-Kx$ the evaluation
equation~\eqref{eq:vn_def} is solved exactly by $v_K(x)=-x^\top P_Kx-c_K$ with
$P_K$ from the Lyapunov equation
\begin{gather*}
  \big(A-BK-\tfrac\lambda2 I\big)^\top P_K
  +P_K\big(A-BK-\tfrac\lambda2 I\big)+Q+K^\top RK=0,
  \\
  c_K=\lambda^{-1}\tr(\sigma\sigma^\top P_K),
\end{gather*}
and $K_{n+1}=R^{-1}B^\top P_{K_n}$ is Kleinman's iteration: it realizes the
ideal sequence $\{v_n\}$ of~\eqref{eq:vn_def} exactly, giving direct access to
the decomposition $\tilde v_n-v_n=\delta_n+\varepsilon_n$ of
Section~\ref{sec:closed} and a measurement of $\rho_{\mathrm{PI}}$ in
Assumption~\ref{ass:ideal_conv}.

\paragraph{Implementation}
$\tilde v_n$ is a plain $\tanh$ MLP with three hidden layers of width $256$
($128$ for the two-dimensional buffer study). Inputs are normalized by
$L_{\mathrm{train}}$, and the output is scaled by a rollout estimate under
the initial policy; both are non-dimensionalizations that never touch the
reference. This normalization is important under the optimization budgets
considered here; without it, training frequently collapses toward a nearly
constant surrogate (\ref{app:supp}).

We deliberately avoid a quadratic ansatz, a linear-feedback policy class, and
any anchoring to the analytic constant $c$, all of which encode knowledge of
the reference; Section~\ref{sec:exp_highdim} and~\ref{app:ansatz} quantify
what they are worth. No anchoring is needed for well-posedness: coercivity
($\lambda>0$) already determines the additive constant.

The initial policy is $\tilde a_0\equiv0$ unless stated otherwise;
Theorem~\ref{thm:wellposed} ensures that this policy has a well-defined PDE
evaluation even though it does not stabilize the plant, and its numerical
cost is examined on
the quadrotor (Section~\ref{sec:exp_nonlinear}). Two deliberate exceptions:
the attenuation probe of
Section~\ref{sec:exp_buffer} freezes the analytic optimal feedback, and the
quadrotor runs start from the clipped
LQR feedback of the hover linearization.

We write $N_{\mathrm{eval}}$ for the optimizer steps per policy evaluation
(the \emph{per-evaluation budget}), $N_{\mathrm{PI}}$ for the number of outer
policy updates, and $N_{\mathrm{tot}}:=N_{\mathrm{eval}}N_{\mathrm{PI}}$ for
the \emph{total optimization budget}. Each run: $8192$ collocation points on
$D_{\mathrm{train}}$, resampled every $25$ steps;
$N_{\mathrm{eval}}=1200$ and $N_{\mathrm{PI}}=10$, so
$N_{\mathrm{tot}}=1.2\times10^4$ ($N_{\mathrm{PI}}=12$ in
Section~\ref{sec:exp_recursion}; the buffer sweep of
Section~\ref{sec:exp_buffer} uses $N_{\mathrm{PI}}=8$,
$N_{\mathrm{eval}}=2000$ with density-matched collocation, stated at
Figure~\ref{fig:buffer_decay}); a single Adam instance at $10^{-3}$,
cosine-annealed over the whole run, since re-creating it per policy update
restarts the schedule and repeatedly displaces the surrogate. The Laplacian
in~\eqref{eq:residual} is exact for $d\le12$ and a Rademacher estimator with
$8$ probes above that; every reported residual is recomputed exactly. All
runs use a single NVIDIA GeForce RTX 4080 SUPER (16\,GB) with PyTorch 2.4.1
and CUDA 12.4, in float32 with TF32 disabled: the $10$-bit TF32 mantissa
floors relative accuracy at $2\times10^{-4}$, whereas float32 reaches
$1.7\times10^{-7}$. The code, together with the measured data behind every
table and figure reported below, is available at
\url{https://github.com/yeoneung/neural_policy_iteration}.

\paragraph{Measured quantities}
Table~\ref{tab:measured} maps each quantity of the analysis to its estimator.
All reported validation norms are Monte Carlo: $2\times10^5$ samples for
values, $5\times10^4$ for gradients, and $6.6\times10^4$ fresh uniform
samples with the exact Laplacian for residuals. For each benchmark dimension
$d$ we use $p=d+1$, the smallest integer exponent satisfying the theoretical
requirement $p>d$, and compute the theorem-aligned residual statistic
$\widehat\tau_n$ from Remark~\ref{rem:empirical_residual}. When a single
residual number is reported for a run it is the final iterate's
$\widehat\tau_{N_{\mathrm{PI}}}$, written $\widehat\tau$. This is an
empirical residual statistic rather than an estimator of the theoretical
supremum $\bar\tau$; empirically, the per-iterate values are nearly
constant once the iteration reaches its floor. Unless stated otherwise, summaries are geometric
means over the three seeds with $95\%$ intervals; the seed draws both
$(A,B)$ and the initialization, and three seeds is a small sample, so the
intervals should be read accordingly. Two deliberate exceptions to this convention:
ranges over seeds are shown where the seed distribution is wide or bimodal
(Table~\ref{tab:scaling}), and the closed-loop gap of
Section~\ref{sec:exp_nonlinear} is reported as mean $\pm$ standard error
because it can reach zero or change sign, where geometric statistics are
undefined.

\begin{table}[htbp]
  \centering
  \caption{Quantities in the analysis and their empirical counterparts. $D$
  denotes the fixed evaluation domain and $D_{\mathrm{train}}$ the training
  domain; the buffer width is
  $\varrho=\operatorname{dist}(D,\partial D_{\mathrm{train}})$.}
  \label{tab:measured}
  \small\setlength{\tabcolsep}{3pt}
  \begin{tabular}{@{}lll@{}}
    \toprule
    Theory & Empirical estimator & Used in \\
    \midrule
    $\|\tilde v_n-V\|_{L^\infty(D)}$ & max abs.\ error vs $V^\star$ on $D$ & all \\
    $\tau_n$ & $\widehat\tau_n$, fresh-sample estimate of $\|r_n\|_{L^p(D_{\mathrm{train}})}$ & \S\S\ref{sec:exp_buffer},\,\ref{sec:exp_recursion},\,\ref{sec:exp_highdim} \\
    $G_k$ & $G_k^{\mathrm{emp}}:=\max|\nabla\tilde v_k-\nabla v_k|$ on $D$, $v_k$ from Kleinman & \S\ref{sec:exp_recursion} \\
    $\gamma$ & $\widehat\gamma$ of~\eqref{eq:gamma_hat}, median ratio in the geometric phase & \S\ref{sec:exp_recursion} \\
    $\theta_{\mathcal D}=B_a/\mu_a$ & $\|B\|/(2\lambda_{\min}(R))$ (control-affine, $L_{ab}=0$) & \S\ref{sec:exp_recursion} \\
    $\kappa_d=\kappa/\sqrt d$ & realized rate~\eqref{eq:realized_rate}, compared with $\kappa$, $\kappa_d$ & \S\ref{sec:exp_buffer} \\
    $\rho_{\mathrm{PI}}$ & decay rate of $\|v_n-V^\star\|_{L^\infty(D)}$ (Kleinman) & \S\ref{sec:exp_recursion} \\
    $\widetilde G_{\mathcal D},\widetilde K$ & $\sup_n\|\nabla\tilde v_n\|_{L^\infty(D_{\mathrm{train}})}$, $\sup_n\|\tilde v_n\|_{L^\infty(D_{\mathrm{train}})}$ & \S\ref{sec:exp_recursion} \\
    \bottomrule
  \end{tabular}
\end{table}

\subsection{Interior attenuation and buffer effects}
\label{sec:exp_buffer}

Lemma~\ref{lem:attenuation} separates a finite-$p$ source term from an
exponentially attenuated homogeneous boundary term for the linear operator
$\mathcal T_0w:=\lambda w-\tfrac12\tr(\sigma\sigma^\top D^2_{xx}w)-b\cdot\nabla_xw$.
The finite-difference probe below isolates the homogeneous term ($r=0$), whose
predicted decay rate is independent of the residual norm. We test that rate on
the operator directly, by finite differences: we solve $\mathcal T_0w=0$ with
unit Dirichlet data $w\equiv1$ on the boundary of the box and measure the
decay of $\max|w|$ against the distance to the boundary. A network
resolving a thin boundary layer would report its own approximation power, not
the equation's attenuation. In one dimension the solve uses central
differences on $[-L,L]$, $L=6$, with $4000$ interior points; on the square
$[-4,4]^2$, central differences for the diffusion and first-order upwinding
for the advection on a $220^2$ grid. The realized rate is
\begin{equation}\label{eq:realized_rate}
\min_{x:\;\operatorname{dist}(x,\partial U)\ge0.25\,d_{\max}}
\;
\frac{-\log\big(|w(x)|/\|w\|_{L^\infty}\big)}{\operatorname{dist}(x,\partial U)},
\qquad
d_{\max}:=\max_{x\in U}\operatorname{dist}(x,\partial U),
\end{equation}
the slowest normalized decay over the deep interior, with no curve fitting.
We then ask whether this mechanism is what makes boundary-free training work.

\paragraph{Sharpness in one dimension and the dimensional factor}
For constant drift the homogeneous solutions of $\mathcal T_0w=0$ are
$e^{r_\pm x}$, and the slower rate
$(-|b|+\sqrt{b^2+2\sigma^2\lambda})/\sigma^2$ is exactly the $\kappa$
of~\eqref{eq:kappa}: the worst case the estimate encodes is a drift carrying
boundary data inward. Finite differences on $[-L,L]$ over $30$ configurations
($\lambda\in\{0.5,2\}$, $\sigma\in\{0.25,0.5,1\}$, $b\in\{-2,\dots,2\}$)
confirm it: the realized rate equals $\kappa$ to three digits for $b\neq0$, and
the bound holds with margin $0.5$--$1.0$. On the square, $12$ configurations
realize $\kappa$ rather than $\kappa_d=\kappa/\sqrt2$ (ratio in
$[0.98,1.02]$): the dimensional loss is an artifact of the coordinatewise
$\cosh$ barrier in the proof, absorbed by the prefactor $2d$.

\begin{figure}[htbp]
  \centering
  \pinnfig{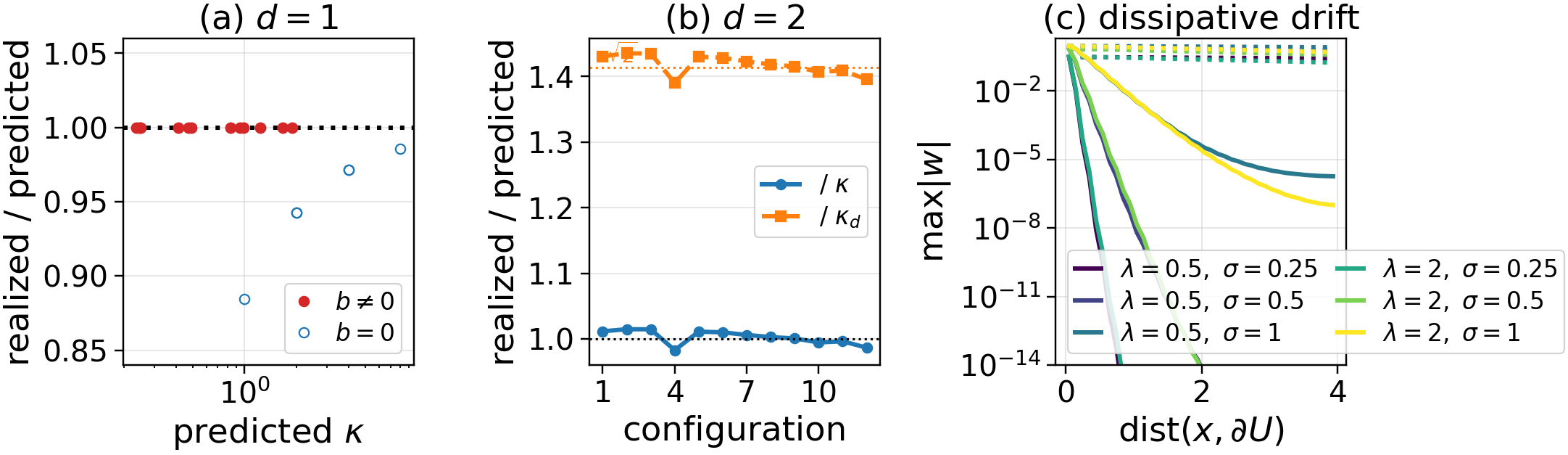}{\textwidth}
  \caption{Sharpness of Lemma~\ref{lem:attenuation}, measured on the operator
  by finite differences with unit Dirichlet data on the box boundary; the
  realized rate is~\eqref{eq:realized_rate}, the predicted rate is the
  $\kappa$ of~\eqref{eq:kappa}.
  (a) $d=1$, constant drift, all $30$ configurations
  ($\lambda\in\{0.5,2\}$, $\sigma\in\{0.25,0.5,1\}$, $b\in\{-2,\dots,2\}$):
  ratio of realized to predicted rate against $\kappa$. Every $b\neq0$
  configuration realizes $\kappa$ exactly ($\pm b$ coincide since $\kappa$
  depends on $|b|$); the $b=0$ rates (open) fall short by
  $\approx\log2/(\kappa L)$ because the symmetric profile contains
  contributions from both boundaries, a shortfall that vanishes as
  $\kappa L$ grows.
  (b) $d=2$, the
  twelve configurations $\lambda\in\{0.5,2\}$, $\sigma\in\{0.25,0.5,1\}$,
  axis-aligned drift $b=(b,0)$ with $b\in\{1,2\}$, ordered by $\kappa$: the
  realized rate is $\kappa$, not
  $\kappa_d=\kappa/\sqrt d$, so the dimensional factor is a proof artifact.
  (c) Dissipative closed loop $b(x)=(A-BK^\star)x$: realized decay (solid) is
  orders of magnitude faster than the bound (dotted); the corresponding
  $\ell_6$ are in Table~\ref{tab:kappa}, and values below $10^{-12}$ are
  solver noise.}
  \label{fig:attenuation}
\end{figure}

\paragraph{Dissipative drift}
The drift relevant to Algorithm~\ref{algo:main} is the closed loop
$b(x)=(A-BK^\star)x$ with $A-BK^\star$ Hurwitz --- the opposite of the worst
case, which the lemma must nonetheless price at the sup-norm bound
$B_{\mathcal D}$. Figure~\ref{fig:attenuation}(c) and Table~\ref{tab:kappa}
report the gap. Writing $B(s)$ for the measured boundary-influence profile
(the maximum of $|w|$ over grid points whose Chebyshev distance to the
boundary falls in the bin at $s$, using $60$ uniform bins), we define
\begin{equation}\label{eq:ell6}
\ell_6:=\inf\{s\ge0:\;B(s)/B(0)\le10^{-6}\}
\end{equation}
and report the first bin center below the threshold, computed on an $800^2$
grid; the bound permits
$\ell_6^{\mathrm{bound}}=\kappa_d^{-1}\log10^6$. The measured $\ell_6$ is
$0.23$--$3.2$ units, against the $86$--$388$ the bound permits, a factor
of $34$ to $1660$. Boundary information therefore decays far faster under a
stabilizing feedback than the worst-case bound permits. The effect is strongly noise
dependent ($\ell_6$ grows tenfold from $\sigma_0=0.25$ to $1$): a low-noise
phenomenon, not a universal one.

At $\lambda=0.5$ and $\sigma_0=0.25$ (the low-noise LQR setting used in
the exact-reference experiments that follow), boundary influence is below
$10^{-6}$ within $0.23$ units, which
is why the buffer sweeps below are flat, and why the Riccati reference is
admissible: the saturation region at $|x|_\infty\approx2.5$ lies many
attenuation lengths from $D$, and a $d=2$ grid solve of the constrained
equation agrees with $V^\star$ on $D$ to $10^{-4}$ relative. That residue is
discretization error rather than a genuine gap: it is independent of the
control box and falls fourfold when the grid is halved.

\begin{figure}[htbp]
  \centering
  \pinnfig{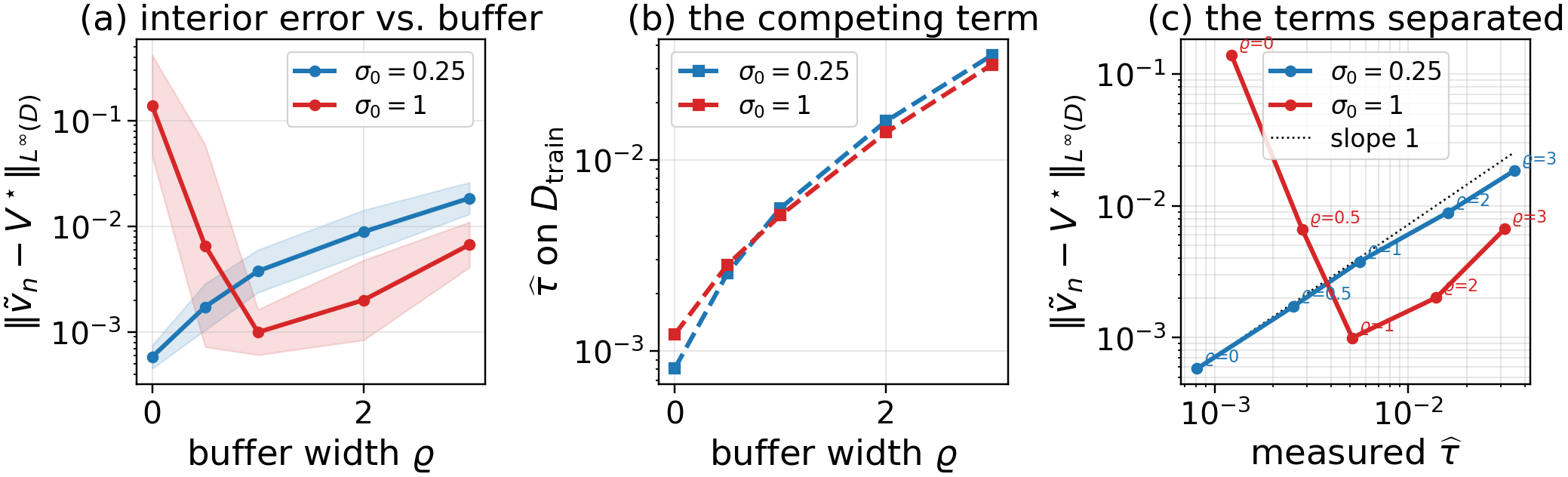}{\textwidth}
  \caption{The buffer prediction test at $d=2$; only the noise level differs
  between the sweeps. Geometry: $D=[-1,1]^2$ fixed,
  $D_{\mathrm{train}}(\varrho)=[-(1+\varrho),1+\varrho]^2$. Each point is the
  final iterate after $N_{\mathrm{PI}}=8$ policy updates at
  $N_{\mathrm{eval}}=2000$, with the collocation density
  matched across $\varrho$ ($4096$ points at $\varrho=0$ up to $65536$ at
  $\varrho=3$). (a) Interior error against buffer width: no benefit
  predicted at $\sigma_0=0.25$, benefit predicted at $\sigma_0=1$; both hold
  (stars: best buffer). (b) The competing term: at a fixed
  optimizer-update budget the
  theorem-aligned residual $\widehat\tau$ grows with $\varrho$.
  (c) Error against measured $\widehat\tau$: the low-noise sweep
  follows the slope-one line, the residual-dominated regime of
  Proposition~\ref{prop:delta_closed}, while the high-noise sweep descends
  onto it as the buffer absorbs the boundary term. Bands: $95\%$ CI over
  three seeds.}
  \label{fig:buffer_decay}
\end{figure}

\paragraph{A prediction and its test}
These are measurements of the equation, not of the method, so they predict the
method's behavior in advance. The finite-difference map of $\ell_6$
(Figure~\ref{fig:regime_map}) says, everything but the noise held fixed:
\[
  \sigma_0=0.25:\;\ell_6=0.23\;\Rightarrow\;\text{no benefit},
  \qquad
  \sigma_0=1:\;\ell_6=3.23\;\Rightarrow\;\text{benefit}.
\]
The PINN sweep (Figure~\ref{fig:buffer_decay}) confirms both predictions:
the low-noise error is best at zero buffer and degrades $32\times$ as the
domain widens, while the high-noise error falls $139\times$ up to
$\varrho=1$ and then turns.
The turn is the competing term of Theorem~\ref{thm:closed}: although the
collocation density is held fixed, the same number of parameter updates
must reduce the residual over an increasingly large domain, and
$\widehat\tau$ grows $26\times$ across the
high-noise sweep. Panel~(c) separates the two terms: against
$\widehat\tau$, the low-noise sweep follows the slope-one line while the
high-noise sweep descends onto it as the buffer absorbs the boundary term.

\paragraph{Consequences for the method}
These experiments separate the two terms of the interior estimate. In the
low-noise dissipative regime the boundary contribution is already
negligible at zero buffer, so enlarging the training domain only increases
the residual error; at higher noise, a moderate buffer first reduces the
boundary contribution (in the $\sigma_0=1$ configuration, one unit of
buffer reduces it by approximately two orders of magnitude), after which
the increasing residual becomes dominant. The attenuation scale can
therefore be measured independently of the neural solver and used to
anticipate when a training buffer is beneficial.

\subsection{Error recursion and empirical contraction}
\label{sec:exp_recursion}

The ideal iterates $v_n$ are available in closed form
(Section~\ref{sec:exp_setup}), so the empirical counterpart
\[
G_k^{\mathrm{emp}}
:=
\|\nabla_x\tilde v_k-\nabla_x v_k\|_{L^\infty(D)}
\]
of the gradient error driving Proposition~\ref{prop:closed_recursion} is
measured, not inferred, on the fixed evaluation domain $D$ rather than on the
nested theoretical domains $D^{(k)}$, which enter only the proof. Both
sequences start from $a_0=\tilde a_0\equiv0$. All runs: $d=5$,
$\lambda=0.5$, three seeds. The observed floor and the measured contraction
are
\begin{equation}\label{eq:gamma_hat}
\begin{aligned}
G_{\mathrm{floor}}
&:=
\operatorname{median}
\big\{
G_{N_{\mathrm{PI}}-2}^{\mathrm{emp}},
G_{N_{\mathrm{PI}}-1}^{\mathrm{emp}},
G_{N_{\mathrm{PI}}}^{\mathrm{emp}}
\big\},
\\
\widehat\gamma
&:=
\operatorname{median}
\Big\{
\tfrac{G_{k+1}^{\mathrm{emp}}}{G_k^{\mathrm{emp}}}
:
G_k^{\mathrm{emp}}>3G_{\mathrm{floor}}
\Big\}:
\end{aligned}
\end{equation}
the floor is the median of the final three iterates rather than the
minimum, which is sensitive to a single downward fluctuation, and the
contraction ratios are restricted to the geometric phase
$G_k^{\mathrm{emp}}>3G_{\mathrm{floor}}$ because the naive
ratio is biased toward one once the iteration reaches its floor.

\paragraph{Geometric decay and the residual floor}
Both predicted features are present. Varying $N_{\mathrm{eval}}$ moves the
theorem-aligned residual $\widehat\tau$ by a factor of $18$ and
$G_{\mathrm{floor}}$ by $19$ in the same direction (Table~\ref{tab:recursion});
regressing $\log G_{\mathrm{floor}}$ on $\log\widehat\tau$ over all
fifteen runs gives slope $1.00$ against a predicted one. The relation holds
across the gap between the continuous $L^p$ residual of the proposition and
the empirical collocation loss the algorithm minimizes
(Remark~\ref{rem:empirical_residual}).

\paragraph{The contraction condition}
With $L_{ab}=0$,
$\theta_{\mathcal D}=B_a/\mu_a=\|B\|_2/(2\lambda_{\min}(R))$, so sweeping $R$
over a factor of $80$ moves $\theta_{\mathcal D}$ from $0.23$ to $18.7$,
across the range where the sufficient condition $\gamma<1$ fails, $\gamma$
being proportional to the explicit greedy-map sensitivity factor
$\theta_{\mathcal D}$ with all else held equal. We test whether this
explicit factor predicts the observed contraction of the neural policy
iteration; varying $R$ can also move other problem-dependent quantities
entering the bound, so the sweep probes the factor the estimate makes
explicit rather than a ceteris-paribus derivative. The
measured contraction does not respond: medians $0.52$--$0.76$ at the default
budget, $0.51$--$0.65$ at $2.5\times$ the budget (which lowers the floor and
lengthens the measurable geometric phase), with no monotone trend in
$\theta_{\mathcal D}$. The floor's apparent doubling across
the sweep at the default budget flattens at the higher one
($2.3\times10^{-2}$ to $1.5\times10^{-2}$): a budget artifact, not a
property of the recursion. The sufficient condition $\gamma<1$ therefore
appears substantially conservative on the tested instances:
$\theta_{\mathcal D}$ is the worst-case Lipschitz constant of the greedy map,
and the realized iteration does not visit the worst case. With one to five
ratios per run surviving the geometric-phase restriction and three seeds,
the exponent is determined only coarsely; the conclusion rests on the
\emph{absence} of a trend in $\theta_{\mathcal D}$, not on the value of
$\widehat\gamma$.

\paragraph{Convergence of the ideal iteration}
The relative error of the exact Kleinman sequence runs
$9.7\times10^{-1},\,2.6\times10^{-1},\,3.1\times10^{-2},\,5.5\times10^{-4},\,
1.4\times10^{-7},\,3.1\times10^{-15}$: quadratic convergence, as for Newton's
method on the Riccati equation. Assumption~\ref{ass:ideal_conv} asks only for a
geometric rate and holds with room to spare, so of the iteration-dependent
terms of Theorem~\ref{thm:closed} the ideal-convergence term
$C_D\rho_{\mathrm{PI}}^{\,n}$ is the first to become negligible.

\begin{figure}[htbp]
  \centering
  \pinnfig{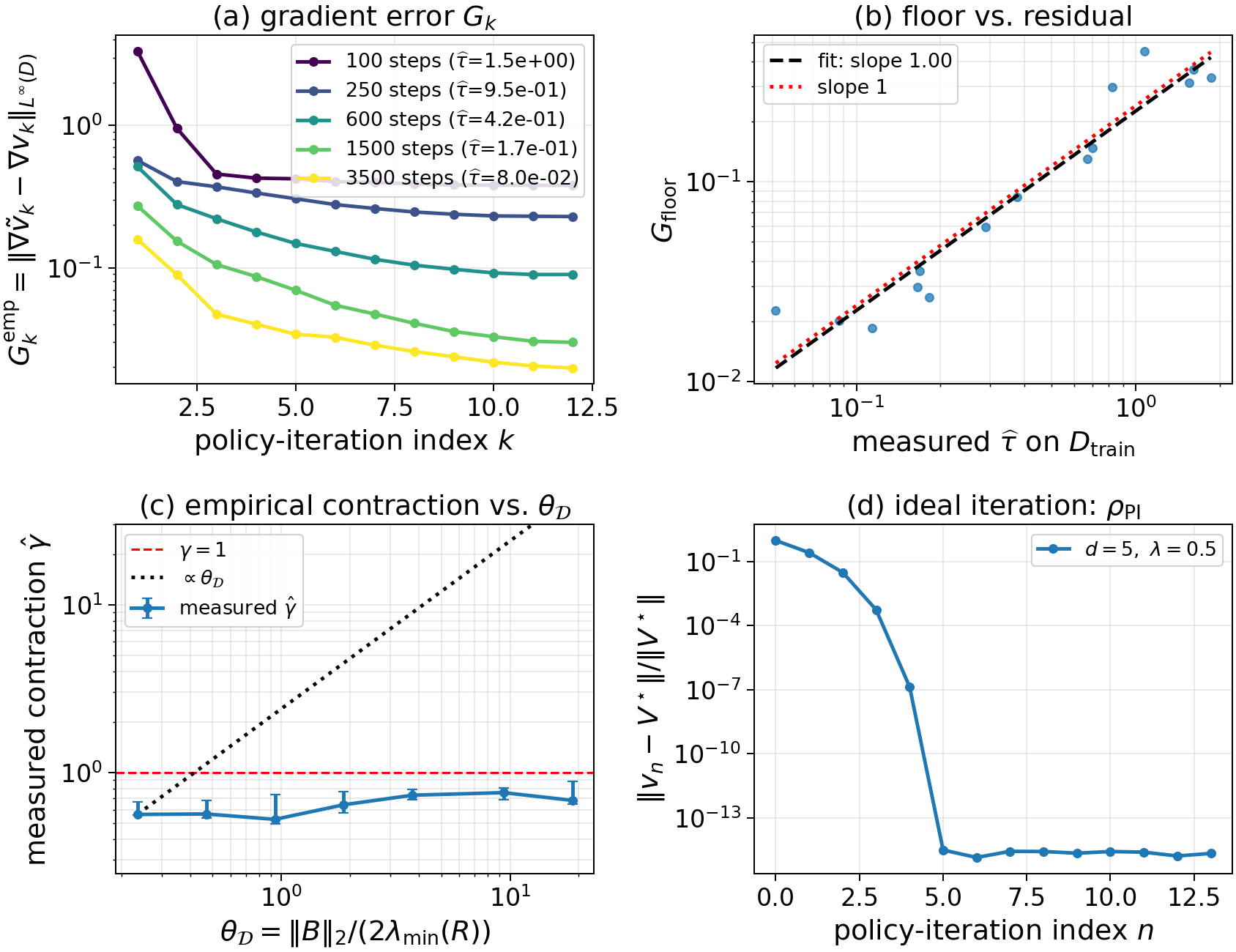}{\textwidth}
  \caption{The closed error recursion of
  Proposition~\ref{prop:closed_recursion} at $d=5$.
  (a) Gradient error $G_k^{\mathrm{emp}}$ against the policy-iteration index
  at five per-evaluation budgets $N_{\mathrm{eval}}$: geometric decay
  followed by a residual-determined floor.
  (b) $G_{\mathrm{floor}}$ against the theorem-aligned measured $L^p$
  residual $\widehat\tau$, log-log; the fitted slope is $1.00$ against a
  predicted one.
  (c) Measured contraction $\widehat\gamma$ of~\eqref{eq:gamma_hat} as
  $\theta_{\mathcal D}=\|B\|_2/(2\lambda_{\min}(R))$ is swept over a factor of
  $80$; the explicit sensitivity factor $\theta_{\mathcal D}$ does not
  predict the observed contraction, which stays below one throughout: the
  sufficient worst-case estimate is conservative.
  (d) The exact ideal (Kleinman) iteration, whose convergence is quadratic, so
  that Assumption~\ref{ass:ideal_conv} holds with room to spare.}
  \label{fig:recursion}
\end{figure}

\subsection{Fixed-policy versus direct HJB training}
\label{sec:exp_pi}

The premise of the method is that a sequence of \emph{linear} fixed-policy
residuals trains better than the nonlinear residual
\[
  \lambda v-\tfrac12\tr(\sigma\sigma^\top D^2_{xx}v)
  -\sup_{a\in A}\big\{b(\cdot,a)\cdot\nabla_xv+L(\cdot,a)\big\}=0,
\]
which is what a deep Galerkin-style solver~\citep{sirignano2018dgm}
minimizes; the two differ only in
where the supremum is taken. Same network, sampler, schedule and step budget for
both; measured wall-clock agrees within $5\%$, so step-matched is time-matched.

\begin{table}[htbp]
  \centering
  \caption{Policy iteration against direct minimization of the nonlinear HJB
  residual at matched budget. Geometric means over three seeds with $95\%$
  intervals; the ratio is the geometric mean of per-seed ratios (above one:
  direct is worse).}
  \label{tab:pi_vs_direct}
  \small\setlength{\tabcolsep}{3pt}
  \begin{tabular}{@{}ccccc@{}}
    \toprule
    $d$ & PINN-PI, rel.\ $L^2$ & direct, rel.\ $L^2$
    & ratio direct/PI & verdict \\
    \midrule
    2  & $6.6\times10^{-3}$ $[5.5,7.9]$ & $5.3\times10^{-3}$ $[2.9,9.9]$
       & $0.81$ $[0.52,1.25]$ & indistinguishable \\
    5  & $1.0\times10^{-2}$ $[0.7,1.4]$ & $2.0\times10^{-1}$ $[0.009,4.5]$
       & $19.6$ $[0.75,510]$ & inconclusive \\
    10 & $3.1\times10^{-1}$ $[2.8,3.4]$ & $8.8$ $[7.2,11]$
       & $28.5$ $[24,34]$ & PINN-PI \\
    \bottomrule
  \end{tabular}
\end{table}

\begin{figure}[htbp]
  \centering
  \pinnfig{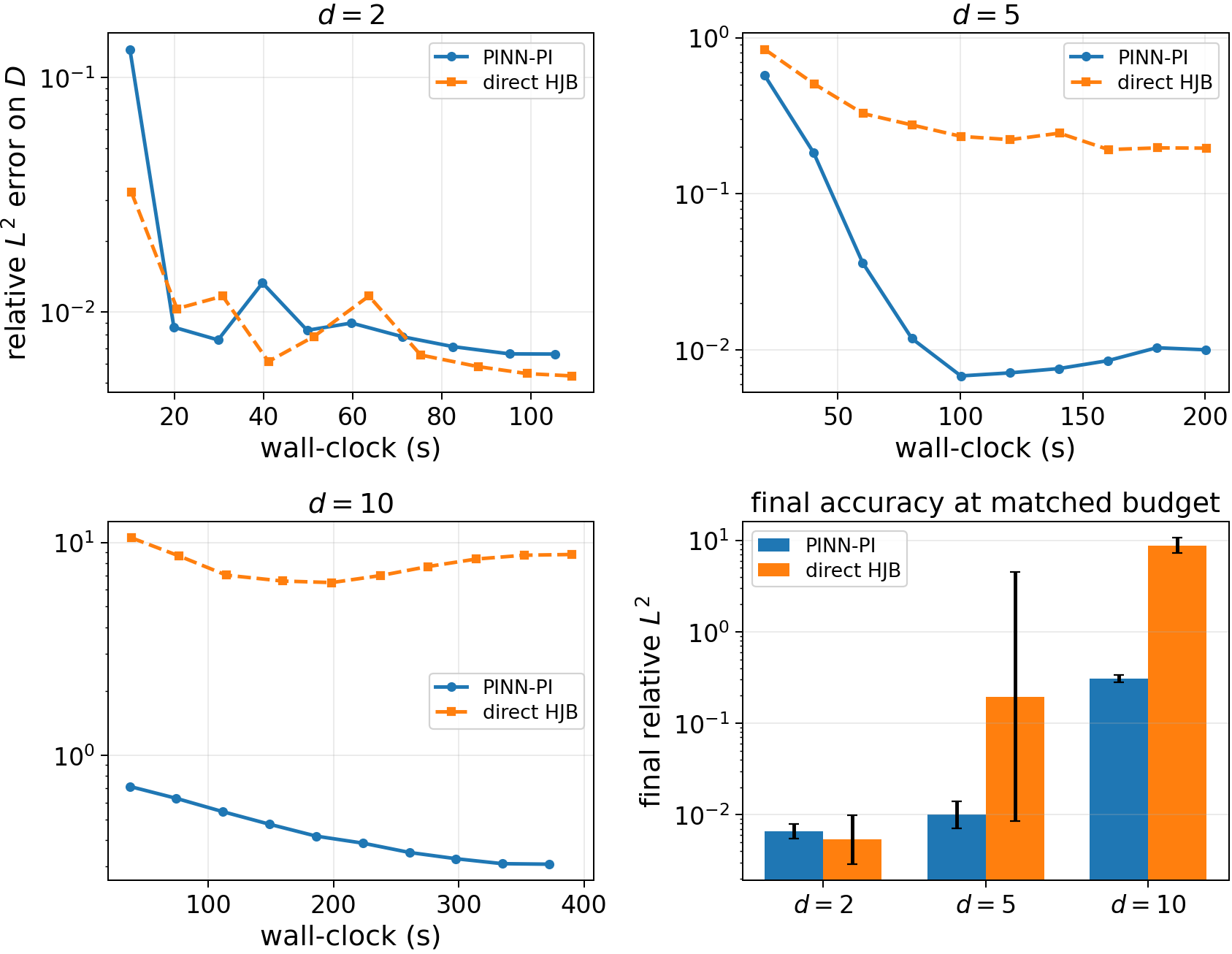}{\textwidth}
  \caption{Policy iteration against direct minimization of the nonlinear HJB
  residual: relative $L^2$ error against wall-clock at $d\in\{2,5,10\}$
  (geometric mean over three seeds) and final accuracy with $95\%$ intervals.
  Same network, sampler and step budget throughout. At $d=5$ the direct
  solver plateaus more than an order of magnitude above policy iteration; at
  $d=10$ its error stops decreasing mid-run while policy iteration is still
  descending.}
  \label{fig:pi_vs_direct}
\end{figure}

The pattern in Table~\ref{tab:pi_vs_direct} and
Figure~\ref{fig:pi_vs_direct} is graded. At $d=2$ the two are
indistinguishable. At $d=5$ the direct solver fails on one seed in three
($4.4$ against $8.3\times10^{-3}$); the failure is intermittent, so the
interval spans one.
At $d=10$ the failure is systematic: all seeds between $24$ and $33$, and the
direct solver at $8.8$ no longer provides an accurate approximation under
the reported error metric. The difference is not explained by residual
magnitude alone: at $d=10$ the direct solver's residual, measured for both
methods on the same nonlinear HJB equation, is only $1.2$--$1.7\times$ larger,
while its value error is $28\times$ larger. Both reduce the residual
comparably but select different small-residual functions. This is the effect of
the linearization: the inner problems of Algorithm~\ref{algo:main} are linear,
while the direct objective passes through a supremum and is not convex in the
surrogate.

Two caveats. Policy iteration is not good at $d=10$ either
($3.1\times10^{-1}$; Section~\ref{sec:exp_highdim} shows why), so the claim is
comparative. And the gap is not an artifact of the function class: with the
quadratic architecture of Section~\ref{sec:exp_highdim} for both methods,
policy iteration reaches $6.1\times10^{-3}$ at $d=10$ against
$6.7\times10^{-1}$ direct (per-seed ratio $61$--$258$, geometric mean $109$),
ratio $5.9$ at $d=5$, parity at $d=2$. This ablation supports attributing
the observed advantage primarily to the training structure rather than to
the hypothesis class.

\subsection{Dimension scaling and computational cost}
\label{sec:exp_highdim}

\paragraph{Accuracy and cost across dimensions}
Table~\ref{tab:scaling} ($d\in\{2,5,10,20,50\}$, fixed total optimization
budget $N_{\mathrm{tot}}=1.2\times10^4$): accuracy is good to $d=5$ and
degrades sharply after.
The increase in the validated residual tracks the observed degradation,
consistent with Proposition~\ref{prop:delta_closed}: the theorem-aligned
residual $\widehat\tau$ rises from $5.6\times10^{-3}$ to $2.8\times10^{1}$,
and the estimate assigns the interior error a floor proportional to it.
Because pointwise gradient control requires $p>d$,
the validated residual moment becomes increasingly tail-sensitive as the
state dimension grows, and $8192$ points cannot control it over $[-2,2]^d$
past $d\approx10$. In these fixed-budget experiments the deterioration is
thus accompanied by a rapid increase in the tail-sensitive finite-$p$
residual, indicating that its control is a dominant limitation; this
observation does not exclude additional approximation or optimization
effects.

\begin{table}[htbp]
  \centering
  \caption{Accuracy and cost against dimension at fixed total budget
  $N_{\mathrm{tot}}=1.2\times10^4$. Geometric
  means over three seeds with ranges; wall-clock from a separate single-job
  pass; memory is the allocator high-water mark. Wall-clock is not monotone
  in $d$ because the Laplacian switches from exact ($d\le12$) to the
  Hutchinson estimator, which is cheaper at $d=20$ than $d$ exact backward
  passes are at $d=10$.}
  \label{tab:scaling}
  \small\setlength{\tabcolsep}{3pt}
  \begin{tabular}{@{}cccccc@{}}
    \toprule
    $d$ & rel.\ $L^2$ & range over seeds & $\widehat\tau$
    & peak memory & wall-clock \\
    \midrule
    2  & $5.7\times10^{-3}$ & $[3.1,9.0]\times10^{-3}$ & $5.6\times10^{-3}$ & \phantom{0}0.7\,GB & \phantom{0}95\,s \\
    5  & $1.2\times10^{-2}$ & $[0.9,1.6]\times10^{-2}$ & $1.3\times10^{-1}$ & \phantom{0}1.5\,GB & 196\,s \\
    10 & $3.1\times10^{-1}$ & $[2.9,3.5]\times10^{-1}$ & $1.3$ & \phantom{0}2.9\,GB & 369\,s \\
    20 & $8.0\times10^{-2}$ & $[1.1\times10^{-2},3.7\times10^{-1}]$ & $6.8$ & \phantom{0}5.6\,GB & 303\,s \\
    50 & $4.3\times10^{-1}$ & $[1.9,6.6]\times10^{-1}$ & $2.8\times10^{1}$ & 14.0\,GB & 305\,s \\
    \bottomrule
  \end{tabular}
\end{table}

Two entries require comment. The $d=20$ seed range spans a factor of $34$, so its
mean is not evidence of improvement over $d=10$. The $d=50$ peak memory,
$14$\,GB, is at device capacity, so budget and collocation batch trade off
directly there.

The $d=10$ entry is a budget statement, not a capability ceiling, and the two
levers the analysis suggests can be measured factorially. Raising the total
budget to $N_{\mathrm{tot}}=4\times10^4$ alone:
$3.1\times10^{-1}\to2.0\times10^{-2}$. Shrinking the
training box to $L_{\mathrm{train}}=1.25$ alone, which
Section~\ref{sec:exp_buffer} justifies since at this noise level a wide
domain brings no benefit while reducing the sampling density:
$\to8.9\times10^{-2}$. Together:
$8.3\times10^{-3}$, a factor of $38$, with the theorem-aligned residual
falling from $1.3$ to $0.12$ alongside, as
Proposition~\ref{prop:delta_closed} requires. The same two
levers at $d=50$ take $4.3\times10^{-1}$ to $1.9\times10^{-1}$ (box alone) and
to $4.0\times10^{-2}$ together, a factor of eleven with no change to the
architecture. The degradation in Table~\ref{tab:scaling} is therefore a fixed-budget statement about controlling an increasingly tail-sensitive finite-$p$ residual with $p>d$, not a ceiling of the method.

\paragraph{The value of prior structure}
A quadratic ansatz $v(x)=x^\top Sx+\mathrm{MLP}(x)+b$ that contains the LQR
solution improves the relative $L^2$ error by a factor of $173$ at $d=50$,
while a linear-feedback policy class alone is worth nothing; anchoring
$v(0)$ to the analytic constant helps two- to fivefold but uses information
from the reference solution during training. All three structural choices are excluded from
every other result in this paper; the ablation is tabulated and discussed
in \ref{app:supp} (Table~\ref{tab:ansatz}).

\subsection{Nonlinear control benchmarks}
\label{sec:exp_nonlinear}

The two benchmarks here leave the LQR testbed, whose control bound is
inactive on the training domain, and serve different purposes. On the
pendulum the torque bound binds on $99\%$ of the domain, the
required swing-up cannot be produced by a locally sensible policy, and a
validated grid solution measures the distance to optimal. On the quadrotor
($d=6$) no grid solution exists and the default configuration fails; an iLQR
check calibrates the quality of the LQR reference.
Section~\ref{sec:exp_posterior} finds the same failure signature at $d=100$. Both problems are control-affine with
additive noise $\sigma_0=0.1$ and $\lambda=0.5$.

\emph{Pendulum} ($d=2$): the standard inverted pendulum with torque limit
$|u|\le2$, quadratic state cost in the wrapped angle and velocity. The measured
greedy control saturates on $99\%$ of the evaluation domain, so the greedy map
is effectively bang-bang, exercising the compact-action machinery of
Assumption~\ref{ass:main}. A reference is
computed by a monotone upwind scheme with Howard iteration on a $512^2$ grid,
periodic in the angle; the policy-update step maximizes the \emph{discrete}
Hamiltonian, without which the iteration chatters. The reference is validated
by a Monte Carlo evaluation of its own feedback, which reproduces the grid
value to $0.6\%$.

\emph{Planar quadrotor} ($d=6$): thrust preloaded at hover, so $u=0$ hovers.
With no grid reference at $d=6$ we calibrate against the clipped LQR feedback
of the hover linearization, and check how good it is: batched iLQR on the
\emph{nonlinear} dynamics, cold- or warm-started, improves on it by only
$3.1\%$ on $D$ and $0.0\%$ on a domain twice as wide (tilts to $53^\circ$).
The iLQR check thus indicates that the clipped linearized-LQR feedback is
near-optimal over the tested region: the benchmark tests the
solver at $d=6$, and the LQR feedback serves as a validated
near-optimal reference rather than merely a baseline.

Closed-loop quality is reported throughout as the \emph{gap closed}, the
fraction of the attainable improvement over doing nothing that a policy
actually delivers:
\begin{equation}\label{eq:gap_closed}
  \mathrm{gap}(\pi)
  =\frac{J(\pi)-J(\pi_0)}{J(\pi_{\mathrm{ref}})-J(\pi_0)},
  \qquad
  J(\pi)=\mathbb E\int_0^\infty e^{-\lambda t}L(X_t,\pi(X_t))\,\de t ,
\end{equation}
estimated by Monte Carlo from $256$ initial states drawn uniformly on $D$.
Here $\pi_0\equiv0$ (which hovers on the quadrotor) and $\pi_{\mathrm{ref}}$
is the grid-optimal feedback for the pendulum and the clipped linearized LQR
for the quadrotor. Thus $\mathrm{gap}=0$ is no better than doing nothing,
$\mathrm{gap}=1$ matches the reference, and $\mathrm{gap}<0$ is actively
harmful. Raw returns are reported where the scale matters. Since
$\pi_{\mathrm{ref}}$ is optimal only for the pendulum, quadrotor values
above one are possible.

\paragraph{Protocol variants}
The nonlinear experiments combine six named variants, fixed here once;
full model specifications are in \ref{app:benchmarks}.
\begin{itemize}
\item\textbf{Base.} Uniform collocation on the base training box. For the
pendulum, with state $x=(\theta,\dot\theta)$,
$D_{\mathrm{train}}^{\mathrm{base}}=[-\pi,\pi]^2$ and the evaluation domain
is $D=[-0.6\pi,0.6\pi]^2$; quadrotor boxes are in the appendix.
\item\textbf{Buffered.} The training box enlarged by a stated factor, the
evaluation domain unchanged: ``$1.5\times$ enlarged box'' means
$D_{\mathrm{train}}=1.5\,D_{\mathrm{train}}^{\mathrm{base}}
=[-1.5\pi,1.5\pi]^2$ (the whole box, not the buffer width). The
velocity-axis variant enlarges only the $\dot\theta$ coordinate,
$[-\pi,\pi]\times[-1.5\pi,1.5\pi]$. For the quadrotor, ``buffered'' enlarges
every coordinate $1.5\times$ except $\zeta=\sin\theta$, which is bounded by
construction (appendix).
\item\textbf{Anchored.} The evaluation loss gains a rollout term,
\begin{equation}\label{eq:anchor_loss}
\mathcal L_{\mathrm{eval}}
=
\mathcal L_{\mathrm{PDE}}
+
\lambda_{\mathrm{anc}}\,
\frac1{N_{\mathrm{anc}}}
\sum_{j=1}^{N_{\mathrm{anc}}}
\big|
\tilde v_n(x_j)-\widehat J^{\,\tilde a_n}(x_j)
\big|^2,
\end{equation}
with $\lambda_{\mathrm{anc}}=1$, $N_{\mathrm{anc}}=256$,
the anchor points $x_j$ redrawn uniformly on $D$ at the start of each
outer iteration and the targets $\widehat J^{\,\tilde a_n}(x_j)$ estimated
by Euler--Maruyama model rollout of the frozen policy: $16$ trajectories
per point, step $\Delta t=5\times10^{-3}$, horizon $T=10/\lambda=20$
($e^{-\lambda T}\approx5\times10^{-5}$), states clipped componentwise to
three times the evaluation box, and a discounted left-endpoint Riemann sum
for the integral. The clipping is a numerical safeguard; strictly, the
targets are therefore truncated, clipped rollout estimates of the
frozen-policy value rather than exact evaluations of $v^{\tilde a_n}$.
\item\textbf{On-policy.} After each policy update, half of the collocation
bank is refreshed with states visited by model rollouts of the
\emph{current} policy.
\item\textbf{Stopped.} Rollout-monitored outer stopping: stop at the first
update whose rollout-estimated return does not improve, i.e.\ early stopping
with patience one, fixed in advance.
\item\textbf{Accepted.} A candidate policy update is retained only when its
rollout-estimated return improves over the current policy's; otherwise the
previous policy is kept and its evaluation continues to refine.
\end{itemize}
All variants use the same neural optimization budget; the rollout-anchored,
on-policy, stopped, and accepted variants incur additional model-rollout
cost on top of it.

\begin{figure}[htbp]
  \centering
  \pinnfig{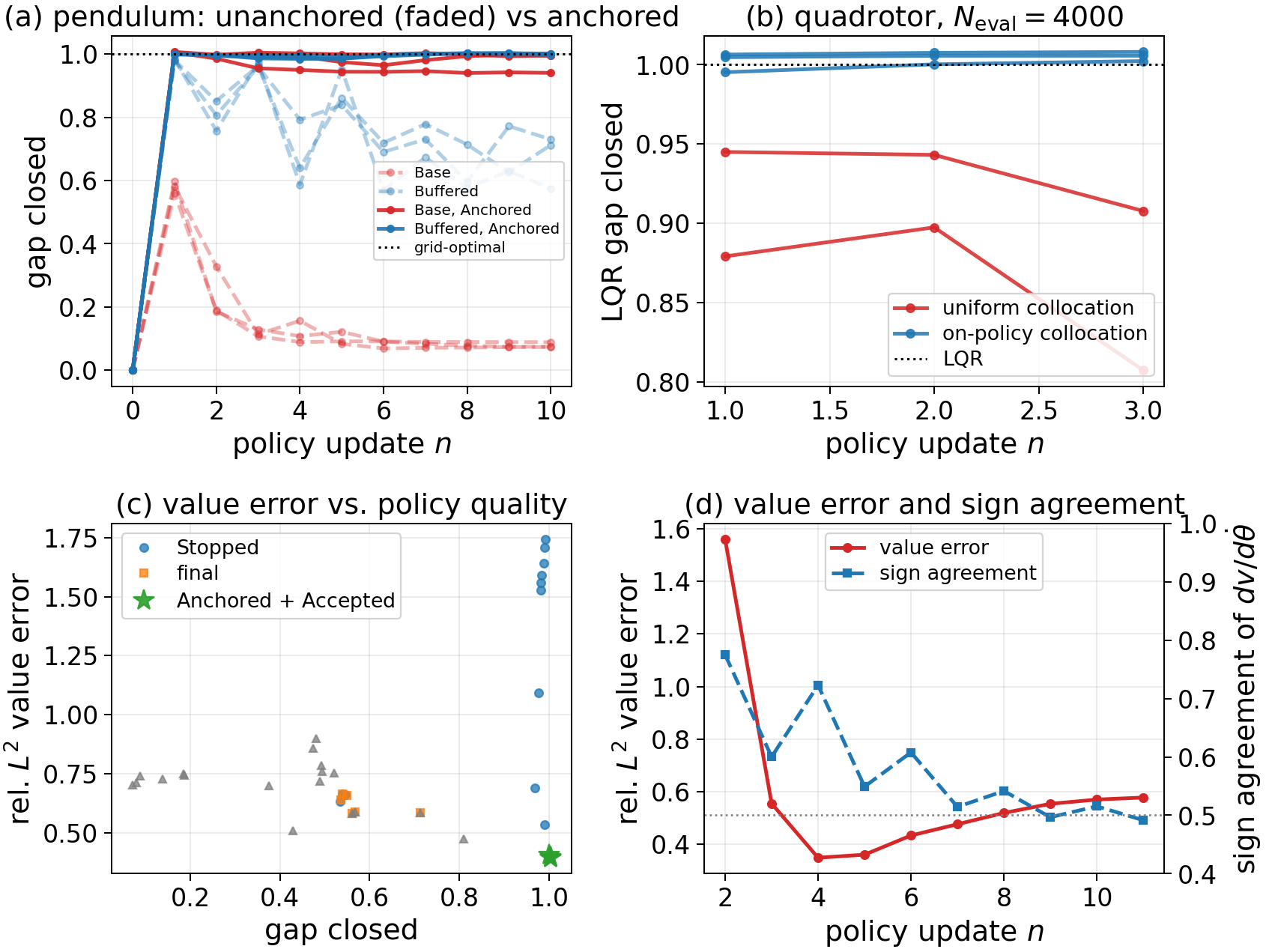}{\textwidth}
  \caption{Per-update closed-loop tracking on the nonlinear benchmarks, three
  seeds each. ``Gap closed'' is~\eqref{eq:gap_closed}: $0$ is no better than
  doing nothing, $1$ matches the reference (grid-optimal for the pendulum,
  clipped linearized LQR for the quadrotor).
  (a) Pendulum, from the zero initial policy ($n=0$, gap $0$ by
  definition): unanchored (faded), the buffer lifts the first-update peak
  from $0.57$ to $0.98$ but the iteration cycles and erodes; anchored, both
  boxes reach gap $\approx1$ at the first update and hold it, with no
  stopping rule.
  (b) Quadrotor at $N_{\mathrm{eval}}=4000$, three updates: uniform
  collocation peaks at the second update and subsequently degrades, whereas
  on-policy collocation reaches and maintains approximately LQR-level
  closed-loop performance.
  (c) Value error
  $E_v:=\|\tilde v-V_{\mathrm{grid}}\|_{L^2(D)}/\|V_{\mathrm{grid}}\|_{L^2(D)}$,
  estimated on $10^5$ uniform samples of $D$ with the $512^2$ grid solution
  bilinearly interpolated, against closed-loop quality, all unanchored
  pendulum runs: the best policies carry the worst value errors, since the
  policy uses only the gradient. Anchored $+$ accept runs (stars) break
  the decoupling (gap $\approx1$ at error $0.40$).
  (d) Unanchored pendulum, per update: $E_v$, and the sign agreement
  $A_{\mathrm{sign}}:=N^{-1}\sum_i
  \mathbf 1\{\operatorname{sgn}\partial_{\dot\theta}\tilde v(x_i)
  =\operatorname{sgn}\partial_{\dot\theta}V_{\mathrm{grid}}(x_i)\}$ over the
  same $10^5$ uniform samples $x_i$ of $D$; this sign is the only quantity
  the bang-bang greedy map uses. The value error is least
  at the fourth update; the sign agreement decays to chance.}
  \label{fig:nonlinear}
\end{figure}

\paragraph{Pendulum: two effects, separated by tracking the iteration}
At the default configuration the surrogate closes only $8\%$ of the gap
between zero control and the grid-optimal feedback. Tracking the closed
loop after \emph{every} policy update (Figure~\ref{fig:nonlinear}(a))
decomposes the failure into two effects. First, the iteration degrades: the
first unanchored update already closes $0.57\pm0.01$ of the gap, but the
subsequent updates enter a period-two degradation. This points to the
evaluation, since Howard iteration under exact evaluation cannot cycle, and
budget and granularity sweeps do not remove it (\ref{app:supp}).
Second, enlarging the training box raises the peak itself, as the
attenuation probe of Section~\ref{sec:exp_buffer} predicts at this noise
level ($\ell_6$ exceeds the domain): the Buffered variant lifts the
first-update peak from $0.57$ to $0.983\pm0.001$ and slows, but does not
remove, the erosion (sampling-geometry diagnostics in \ref{app:supp}).

\paragraph{Anchored evaluation removes the degradation}
The cycle can be traced to the undetermined component of the evaluation.
Trained without boundary data, the evaluation
determines the surrogate only up to what the missing boundary information
leaves free: an additive component and the gradient near the box faces.
The greedy update depends effectively only on the sign of
$\partial\tilde v/\partial\dot\theta$: where that sign is wrong the
bang-bang control flips wholesale, the next evaluation responds to the
flipped policy, and the iteration oscillates between two policies.

The remedy is to determine what the boundary leaves free: the Anchored
variant~\eqref{eq:anchor_loss}. The effect is immediate
(Figure~\ref{fig:nonlinear}(a)): on the buffered box the first anchored
update closes $1.000\pm0.001$ of the gap, no later iterate falls below
$0.98$ over twenty updates, and the feedback swings the pendulum up
on $100\%$ of noisy trajectories at every seed, matching the grid feedback
(the torque bound is $2.0$ against a gravity torque
scale of $15$, so swing-up requires genuine energy pumping). The anchor also
removes, empirically and for this benchmark, the need for an additional
buffer: on the base box the first anchored update
closes $1.006\pm0.001$ and the worst later iterate on any seed is $0.94$,
with swing-up again $100\%$.

Alternatives aimed at the update rather than the evaluation fail to remove
the cycle (\ref{app:supp}), confirming the diagnosis. Combined with the
anchor, however,
the Accepted variant enforces monotonicity with respect to the
rollout-based acceptance metric
($0.999\pm0.003$ at the twentieth update) and drops the value error at the
selected iterate from $1.5$ to $0.40$. The sharp decoupling of value error
from closed-loop quality was itself an artifact of the undetermined
component: the best unanchored policies carried the \emph{worst} value
errors, since the policy uses only $\nabla\tilde v$ while the weakly
determined additive constant dominates the value metric
(Figure~\ref{fig:nonlinear}(c)).

The comparison of Section~\ref{sec:exp_pi}, run where the Hamiltonian is
genuinely nonlinear and the constraint binds: anchored policy iteration
reaches $1.00$ at the last update and swings up, while direct minimization,
which has no frozen policy to anchor and no iteration to stop, reaches
$0.541\pm0.004$ and never swings up (Table~\ref{tab:nonlinear}). The
advantage of fixed-policy training thus becomes more pronounced on the more
difficult tested instances, sharpening the pattern of
Table~\ref{tab:pi_vs_direct}.

\paragraph{Quadrotor: collocation--closed-loop mismatch and on-policy
collocation}
At the default configuration ($\tilde a_0\equiv0$, base training
box) the learned feedback falls below the zero-control baseline: return
$-10.2$ against $-6.9$ for hover ($-1.3$ for the clipped LQR).
Warm-starting at the LQR feedback, the classical admissible
initialization computed from the model alone, eliminates this severe
failure of the zero-policy initialization
but gains little ($0.08$); a buffered box does not help ($-0.57$), so the
failure is not boundary-dominated.
But a \emph{single} policy-evaluation solve at $N_{\mathrm{eval}}=4000$ already
closes $0.88$.

At this budget the iteration degrades: the closed loop peaks at the first
or second update ($0.92\pm0.02$) and falls over ten updates to $0.56$ at
$N_{\mathrm{eval}}=4000$ and $0.03$ at $1200$: the failure of contraction
that Proposition~\ref{prop:closed_recursion} excludes by assumption. We therefore stop on a rollout estimate of the
closed-loop return, at the first update that does not improve it: early
stopping with patience one, fixed in advance. It selects the maximum at
every seed of both benchmarks.

The greedy maps of the two benchmarks differ in kind. The pendulum's
control saturates on $99\%$ of the domain,
so the greedy update depends effectively only on the sign of
$\partial v/\partial\dot\theta$; the relative $L^2$ gradient error stays at
$0.93$--$1.0$ even at the iterate closing $98\%$ of the gap.
Figure~\ref{fig:nonlinear}(d) shows the unanchored
iteration eroding exactly that sign: agreement with the grid reference
falls from $0.78$ at the first update to $0.49$, i.e.\ chance, by the tenth. The quadrotor saturates on under a
quarter of its domain, where the greedy map is smooth in $\nabla\tilde v$
and a larger per-evaluation budget improves rather than degrades the
closed loop.

\paragraph{Diagnosis of the failure}
The learned policy's return ($-10.2$) lies below even saturated random
control ($-8.1$); the control cost is low ($R=0.1$, so the greedy map carries
a gain of $1/2R=5$ into a bound of $3$), which lets a wrong gradient
produce a large wrong control. The learned control is nearly uncorrelated
with the LQR reference on the original training distribution (cosine
$-0.08$ on the collocation, $-0.04$ on $D$) but becomes strongly
anti-aligned along its own closed-loop trajectories ($-0.85$, with no state
in the sample where the two agree in direction): the learned policy drives
the state into a region in which the surrogate gradient is poorly
resolved. Refreshing the collocation set with visited states directly
targets this mismatch, and the $d=100$ posterior problem of
Section~\ref{sec:exp_posterior} exhibits the same signature ($-0.85$ along
controlled trajectories, $+0.11$ on the collocation corridor).

This behavior is
consistent with the scope of Theorem~\ref{thm:closed}: the deterministic
estimate requires residual and gradient control on the dynamically relevant
spatial region, while a small empirical residual on the original
collocation distribution provides no such guarantee, since the empirical
loss may undersample the portion of that region visited by the learned
closed loop. Concretely, the failing
surrogate has sampled maximum HJB residual $0.2$ on the fresh diagnostic
set while the near-optimal LQR value
function has $29$ on the same box: a small residual does not certify the
policy.

\paragraph{Resolution}
The diagnosis suggests the corresponding remedy, which we test against two alternatives
at the same budget (Table~\ref{tab:quad_fixes}). Refreshing half the
collocation along the current policy's own trajectories after each update,
using model rollouts only as in Section~\ref{sec:exp_posterior},
raises the gap closed from $0.921$ to $1.005$, at and slightly past the
clipped LQR feedback, which the iLQR check places within $3\%$ of optimal
(Figure~\ref{fig:nonlinear}(b)).
The iteration also stops degrading: peak and final value coincide.

The two remedies are complementary and the order matters. On-policy
collocation from $\tilde a_0\equiv0$, without the warm start, closes $0.01$
of the gap and the sampled maximum residual rises to $10^2$--$10^3$ (against
$1.35$ with it): under $\tilde a_0\equiv0$ the vehicle tips over, so
sampling along the closed loop concentrates collocation exactly where the
evaluation equation is hardest. A stabilizing $\tilde a_0$ is what makes
the visited states worth sampling.

Alternative remedies aimed at the value function's anisotropy
(per-coordinate input rescaling, curvature whitening) or at sampling
density alone do not remove the shortfall, while tripling the
per-evaluation budget does at three times the cost; the comparison is
tabulated in \ref{app:supp} (Table~\ref{tab:quad_fixes}). Only the remedy
aimed at the sampling distribution removes both the shortfall and the
degradation at unchanged budget.

\begin{table}[htbp]
  \centering
  \caption{Nonlinear benchmarks: gap closed~\eqref{eq:gap_closed}, mean over
  three seeds with s.e., under the named protocol variants of
  Section~\ref{sec:exp_nonlinear}. Pendulum against the grid-optimal
  feedback (return
  $-4.8$; zero control $-11.1$); quadrotor against the clipped linearized LQR
  ($-1.3$; hover $-6.9$). Anchored rows report the \emph{last} update (tenth;
  twentieth with accept), with no stopping rule. Direct minimization has
  neither an iteration to stop nor a policy to anchor.}
  \label{tab:nonlinear}
  \begin{tabular}{@{}llcc@{}}
    \toprule
    Task & variant & PINN-PI & direct \\
    \midrule
    pendulum & Base ($D_{\mathrm{train}}=[-\pi,\pi]^2$) & $0.080\pm0.005$ & $0.004\pm0.013$ \\
    pendulum & Buffered ($1.5\times$ enlarged box) & $0.614\pm0.050$ & $0.541\pm0.004$ \\
    pendulum & Buffered, Stopped & $0.983\pm0.001$ & --- \\
    pendulum & Base, Anchored & $0.977\pm0.026$ & --- \\
    pendulum & Buffered, Anchored & $1.000\pm0.002$ & --- \\
    pendulum & Buffered, Anchored, Accepted & $0.999\pm0.003$ & --- \\
    quadrotor & Base, $\tilde a_0\equiv0$ & $-0.59$ & $-0.60$ \\
    quadrotor & LQR init, Buffered, Stopped & $0.92\pm0.02$ & --- \\
    quadrotor & as above $+$ On-policy & $1.005$ & --- \\
    \bottomrule
  \end{tabular}
\end{table}

\subsection{High-dimensional posterior seeking}
\label{sec:exp_posterior}

As an application outside classical control, we consider finite-time discovery
of low-energy regions of a Bayesian posterior by controlled Langevin dynamics.
We use Bayesian logistic regression for MNIST 3-vs-8 classification with a
$100$-dimensional random Fourier feature map $\phi(x_i)$, labels
$y_i\in\{-1,+1\}$, and a Gaussian prior
$p(\theta)=\mathcal N(0,\sigma_0^2I)$ on $\theta\in\R^{100}$, giving the
negative log-posterior energy
\[
  U(\theta)
  =\sum_i\log\big(1+\exp(-y_i\theta^\top\phi(x_i))\big)
  +\frac{\|\theta\|^2}{2\sigma_0^2},
\]
and control the dynamics
\[
  \de\theta_t=\big(-\nabla U(\theta_t)+Bu_t\big)\,\de t
  +\sqrt{2\sigma_{\mathrm L}}\,\de W_t,
  \qquad\sigma_{\mathrm L}=1.0 .
\]
The reward is $-U(\theta)-\tfrac\alpha2|u|^2$ with $\alpha=0.1$,
$u\in[-10,10]^{16}$, $B\in\R^{100\times16}$ with orthonormal random columns,
and $\lambda=1$; the control is low-rank by design, and its cost is sized so
that a wrong control cannot overpower the drift. The goal is finite-time
discovery of high-probability regions, not exact posterior sampling. Starts
are overdispersed, $\theta_0\sim\mathcal N(0,25I)$, with initial energy
$\sim1.9\times10^4$ against a posterior median of $136$. We report,
for energy thresholds at posterior quantiles $q\in\{0.8,0.5,0.3\}$ (estimated
from a reference chain) and horizons $H\in\{0.5,1,3,5,10\}$, the hit rate and
the conditional mean hitting time, against two baselines: uncontrolled
Langevin, and a random control of the same magnitude.

\paragraph{Failure modes and remedies at $d=100$}
Collocation drawn from the initial law concentrates
at radius $\sigma\sqrt d\approx50$ while the target region sits at radius
$\approx5$, so the value function is unresolved exactly where the
trajectories must go: the collocation--closed-loop mismatch of
Section~\ref{sec:exp_nonlinear} in extreme form. We therefore collocate
half on states visited by Langevin paths, and
refresh part of that bank each policy update with the \emph{current} policy's
own trajectories. A relative residual weighting $(1+U)^{-1}$ is
additionally required because the running cost spans three orders of
magnitude: without it the loss is dominated by the start region, and the
resulting gradient error at the bottom of the landscape creates a trap with
the anti-alignment signature of Section~\ref{sec:exp_nonlinear}. Both
ingredients are individually necessary: with either removed, the
hit rate is zero at every threshold and horizon. These runs predate the
anchored evaluation; three policy updates at
$N_{\mathrm{eval}}=4000$ are used, over which the short-horizon hit rate improves or
holds on every seed, and the myopic descent reference
$u=\mathrm{clip}(-B^\top\nabla U/\alpha)$ calibrates what is achievable
(hit rate $0.96$ at $q=0.8$, $H=3$, against $0.85$ uncontrolled).

\paragraph{Result}
Averaged over three seeds (Table~\ref{tab:posterior}): at $H=3$ the
controlled dynamics reaches the $q=0.8$ region at rate $0.92$ against $0.84$
uncontrolled, and the advantage \emph{grows} with depth ($0.84$ against
$0.61$ at $q=0.5$, $0.73$ against $0.43$ at $q=0.3$), with mean hitting
times uniformly $15$--$20\%$ shorter. The random-control baseline is
indistinguishable from the uncontrolled dynamics, indicating that the
improvement is attributable to the learned feedback rather than to control
magnitude alone. Direct minimization of the
nonlinear HJB residual, given the identical collocation machinery, weighting
and budget, achieves zero hit rate at every threshold, horizon, and seed:
the advantage of fixed-policy training
(Section~\ref{sec:exp_pi}) at its most pronounced.
Two limitations should be noted: at long
horizons uncontrolled Langevin also arrives (the benefit is the transient,
which suggests using the controller as a burn-in phase); and about $3\%$ of
controlled trajectories retain a residual trap tail and do not hit by $H=10$.

\begin{table}[htbp]
  \centering
  \caption{Posterior seeking in $d=100$: hit rate at horizon $H=3$ and mean
  hitting time (over trajectories that hit by $H=10$), mean over three seeds,
  $256$ overdispersed starts each. The myopic reference row is the model-derived
  calibration $u=\mathrm{clip}(-B^\top\nabla U/\alpha)$; the direct solver's
  hitting times are undefined because no trajectory ever hits.}
  \label{tab:posterior}
  \small\setlength{\tabcolsep}{3pt}
  \begin{tabular}{@{}lcccccc@{}}
    \toprule
    & \multicolumn{3}{c}{hit rate at $H=3$} & \multicolumn{3}{c}{mean hitting time} \\
    \cmidrule(lr){2-4}\cmidrule(lr){5-7}
    & $q=0.8$ & $q=0.5$ & $q=0.3$ & $q=0.8$ & $q=0.5$ & $q=0.3$ \\
    \midrule
    PINN-PI      & $0.92$ & $0.84$ & $0.73$ & $2.10$ & $2.38$ & $2.65$ \\
    direct minimization & $0.00$ & $0.00$ & $0.00$ & --- & --- & --- \\
    uncontrolled & $0.84$ & $0.61$ & $0.43$ & $2.45$ & $2.92$ & $3.35$ \\
    random control & $0.83$ & $0.61$ & $0.42$ & $2.46$ & $2.94$ & $3.37$ \\
    myopic reference & $0.96$ & $0.90$ & $0.80$ & $2.01$ & $2.28$ & $2.54$ \\
    \bottomrule
  \end{tabular}
\end{table}

\section{Discussion and Conclusion}

We introduced PINN-PI, a mesh-free policy-iteration method for stationary
second-order HJB equations. The method preserves Howard iteration's structural
advantage: each neural policy-evaluation problem is linear in the value, while
the HJB nonlinearity is handled by a pointwise greedy update.

The main theoretical contribution is an interior analysis for bounded-domain
training without prescribed boundary values. Fixed-policy evaluations are well
posed for every Borel Markov policy in a Lyapunov growth class. A Lipschitz
estimate transfers value-gradient errors to policy errors, and an exponential
attenuation estimate controls unresolved boundary information. On nested
domains, these ingredients yield a closed finite-step recursion with geometric
iteration terms and residual- and amplitude-dependent floors. The result is
purposefully limited: it assumes continuous $L^p$ residual control for a finite exponent $p>d$, an a
priori network bound, and a sufficient feedback-contraction condition; it does
not identify empirical collocation loss with the continuous $L^p$ residual.

The experiments support the mechanisms where they can be measured.
Finite-difference probes recover the sharp one-dimensional attenuation rate and
show that dissipative closed loops attenuate much faster than the worst-case
bound. On the exact-reference LQR problem, the gradient-error floor scales
nearly linearly with the theorem-aligned finite-$p$ residual measured on
fresh validation samples. At matched architecture and
budget, fixed-policy training becomes markedly more reliable than direct
nonlinear HJB residual minimization as the tested problems become more
difficult. Increased
optimization and a smaller training domain substantially improve the
fixed-budget results up to $d=50$.

The nonlinear examples reveal two complementary practical limitations. A
small sampled residual does not certify the policy when collocation points
miss the region visited by the learned closed loop: on the quadrotor and the
100-dimensional Langevin problem, the surrogate gradient becomes
systematically misaligned along its own trajectories despite benign residuals
on the original sample, and trajectory-adapted collocation removes the
misalignment. And
evaluation without boundary data can leave the surrogate free in exactly the
component on which a saturated greedy map depends: on the pendulum the iteration then
cycles instead of converging, and anchoring each evaluation at
rollout-estimated values of its own frozen policy removes the effect, with
the closed-loop gap reaching $1.00$ at the first update and holding it
across twenty while restoring the value metric alongside. These safeguards
(trajectory-adapted collocation, rollout-anchored evaluation, and
rollout-monitored outer stopping) use the model only; they
address the empirical sampling gap and are not formal consequences of the
present finite-$p$ theorem.

The method assumes known dynamics, and the error bound is finite-step on a
fixed domain with nonexplicit contraction constants. Natural extensions
include statistical control of the collocation-to-continuous $L^p$ residual gap,
dissipativity-adapted barriers, expanding-domain analysis with uniform
constants, approximate greedy updates, and learned-dynamics variants.

\section*{Code and data availability}
The implementation, the experiment scripts, and the measured data behind every
table and figure of Section~\ref{sec:experiments} are publicly available at
\url{https://github.com/yeoneung/neural_policy_iteration}.

\appendix
\makeatletter
\@addtoreset{figure}{section}
\@addtoreset{table}{section}
\makeatother
\renewcommand{\thesection}{Appendix~\Alph{section}}
\renewcommand{\thefigure}{\Alph{section}.\arabic{figure}}
\renewcommand{\thetable}{\Alph{section}.\arabic{table}}

\section{Nonlinear benchmark specifications}
\label{app:benchmarks}

Both benchmarks are control-affine diffusions
\[
\de X_t=\big(f(X_t)+G(X_t)u_t\big)\,\de t+\sigma_0\,\de W_t,
\qquad
L(x,u)=-q(x)-u^\top Ru,
\]
with isotropic noise $\sigma_0=0.1$, discount $\lambda=0.5$, and the
pointwise greedy map
$a^\star(x,z)=\operatorname{clip}\big(\tfrac12R^{-1}G(x)^\top z,\,
[-u_{\max},u_{\max}]^m\big)$. Boxes are anisotropic: a box of half-width $h$
means $\prod_i[-s_ih,s_ih]$ for the stated scale vector $s$; the evaluation
domain always uses $h=0.6$, the base training box $h=1$.

\paragraph{Inverted pendulum ($d=2$)}
State $x=(\theta,\dot\theta)$, control $m=1$:
\[
f(x)=\Big(\dot\theta,\;\frac{3g}{2l}\sin\theta\Big),
\qquad
G=\Big(0,\;\frac{3}{ml^2}\Big)^{\!\top},
\qquad
g=10,\quad m=l=1,
\]
so the gravity torque scale is $15$ and the control gain $3$, with torque
bound $|u|\le2$. Running cost
$q(x)=\operatorname{wrap}(\theta)^2+0.1\,\dot\theta^2$ with
$\operatorname{wrap}(\theta)=\operatorname{atan2}(\sin\theta,\cos\theta)$,
and $R=10^{-3}$. Scale $s=(\pi,\pi)$:
$D_{\mathrm{train}}^{\mathrm{base}}=[-\pi,\pi]^2$,
$D=[-0.6\pi,0.6\pi]^2$. Initial policy $\tilde a_0\equiv0$. The grid
reference solves the constrained HJB equation by a monotone upwind scheme
with Howard iteration on a $512^2$ grid, $\theta$ periodic on $[-\pi,\pi)$,
$\dot\theta\in[-8,8]$, $u\in[-2,2]$.

\paragraph{Planar quadrotor ($d=6$)}
State $x=(p_x,p_y,\zeta,v_x,v_y,\dot\zeta)$ with $\zeta=\sin\theta$ encoding
the attitude, so $|\zeta|\le1$ by construction, and
$\cos\theta:=\sqrt{1-\zeta^2}$ (clamped away from zero). Control
$u=(u_1,u_2)$ is (thrust deviation from hover, torque), $m=2$:
\[
f(x)=\Big(v_x,\;v_y,\;\dot\zeta,\;-g\zeta,\;g(\cos\theta-1),\;
-\frac{\zeta\dot\zeta^2}{1-\zeta^2}\Big),
\]
\[
G_{41}=-\frac{\zeta}{m},\quad
G_{51}=\frac{\cos\theta}{m},\quad
G_{62}=\frac{\cos\theta}{I},
\]
all other entries of $G$ zero; $g=9.81$, $m=I=1$, $|u_i|\le3$. Gravity is
cancelled exactly at $\zeta=0$, so $u=0$ hovers. Running cost
$q(x)=\sum_iq_ix_i^2$ with $q=(1,1,1,0.1,0.1,0.1)$ and $R=0.1\,I_2$. Scale
$s=(1,1,0.8,1,1,1)$; the buffered variant multiplies the training
half-widths by $(1.5,1.5,1,1.5,1.5,1.5)$, i.e.\ every coordinate except the
bounded $\zeta$. The LQR warm start linearizes at hover (nonzero entries
$A_{14}=A_{25}=A_{36}=1$, $A_{43}=-g$, $B_{51}=1/m$, $B_{62}=1/I$) and
solves the continuous algebraic Riccati equation for the discount-shifted
pair $(A-\tfrac\lambda2I,\,B)$ with $Q=\operatorname{diag}(q)$ and $R$,
giving $K=R^{-1}B^\top P$ and the initial policy
$\tilde a_0(x)=\operatorname{clip}(-Kx,\,[-3,3]^2)$.

The $d=100$ posterior-seeking problem is fully specified in
Section~\ref{sec:exp_posterior}.

\section{Supplementary diagnostics and ablations}
\label{app:supp}

\subsection{Output-scale calibration}

Adam moves parameters by roughly the learning rate per step, so without
output scaling the surrogate cannot reach an output of $O(10)$ within
budget and collapses to the best constant $\mathbb E[L]/\lambda$. The crude
bound $\sup|L|/\lambda$ overshoots by two orders of magnitude at $d=50$,
whereas the rollout estimate under the initial policy stays within a factor
$1.6$ of the true scale at every dimension tested.

\subsection{Attenuation regime map and dissipative closed-loop rates}

\begin{figure}[htbp]
  \centering
  \pinnfig{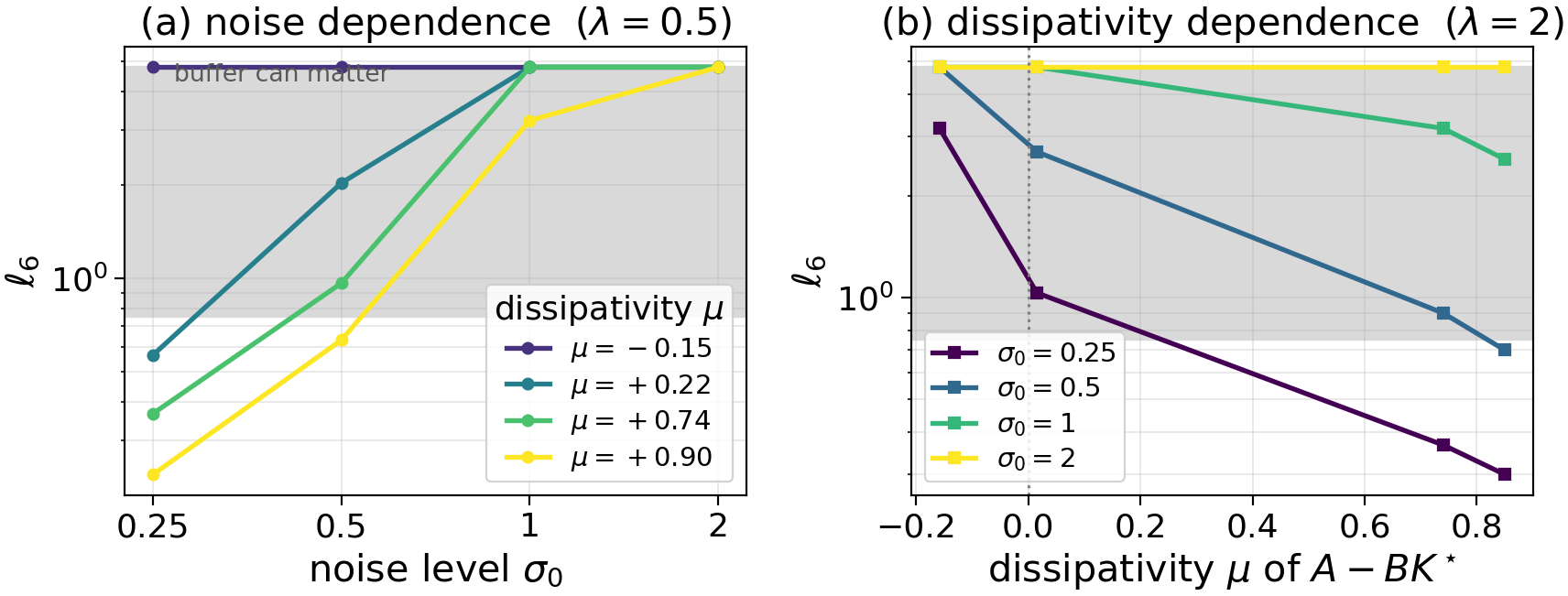}{0.92\textwidth}
  \caption{Where a buffer can matter, from the equation alone: $\ell_6$ against
  the noise level (a) and the closed-loop dissipativity
  $\mu=-\lambda_{\max}\big((A_{\mathrm{cl}}+A_{\mathrm{cl}}^\top)/2\big)$,
  $A_{\mathrm{cl}}:=A-BK^\star$, (b).
  The shaded band marks $\ell_6$ comparable to the training margin; the
  benchmarks of Sections~\ref{sec:exp_recursion}--\ref{sec:exp_highdim} sit
  well below it.}
  \label{fig:regime_map}
\end{figure}

Figure~\ref{fig:regime_map} maps, from the equation alone, where a training
buffer can matter; Table~\ref{tab:kappa} lists the measured $\ell_6$ for the
dissipative closed loop against the values the bound permits.

\begin{table}[htbp]
  \centering
  \caption{Attenuation for the dissipative closed loop $b(x)=(A-BK^\star)x$ on
  $[-4,4]^2$, by finite differences. $\ell_6$ is defined
  in~\eqref{eq:ell6}, which is unambiguous unlike a fitted rate
  since the profile is not a pure exponential; the bound permits
  $\ell_6^{\mathrm{bound}}=\kappa_d^{-1}\log10^{6}$. Grid $800^2$, all cell
  P\'eclet numbers below $1.9$, $\ell_6$ converged from above, so the values
  are conservative.}
  \label{tab:kappa}
  \begin{tabular}{@{}cccccc@{}}
    \toprule
    $\lambda$ & $\sigma_0$ & $\kappa_d$ & $\ell_6^{\mathrm{bound}}$
    & $\ell_6$ (measured) & $\ell_6^{\mathrm{bound}}/\ell_6$ \\
    \midrule
    0.5 & 0.25 & 0.0357 & 387.0 & 0.233 & 1660 \\
    0.5 & 0.5  & 0.0357 & 387.1 & 0.633 &  612 \\
    0.5 & 1    & 0.0356 & 387.9 & 3.229 &  120 \\
    2   & 0.25 & 0.1604 &  86.1 & 0.300 &  288 \\
    2   & 0.5  & 0.1600 &  86.4 & 0.699 &  124 \\
    2   & 1    & 0.1585 &  87.2 & 2.563 &   34 \\
    \bottomrule
  \end{tabular}
\end{table}

\subsection{Residual-floor values}

Table~\ref{tab:recursion} lists the values plotted in
Figure~\ref{fig:recursion}(b).

\begin{table}[htbp]
  \centering
  \caption{The observed floor tracks the $L^p$ residual, with
  $G_{\mathrm{floor}}$ as in~\eqref{eq:gamma_hat}. Geometric means over
  three seeds; the regression of $G_{\mathrm{floor}}$ on $\widehat\tau$ over
  all runs has slope $1.00$ against a predicted one.}
  \label{tab:recursion}
  \begin{tabular}{@{}cccc@{}}
    \toprule
    $N_{\mathrm{eval}}$ & $\widehat\tau$ & $G_{\mathrm{floor}}$
    & $G_{\mathrm{floor}}/\,\widehat\tau$ \\
    \midrule
    \phantom{0}100  & $1.5$                & $3.8\times10^{-1}$ & $0.26$ \\
    \phantom{0}250  & $9.5\times10^{-1}$   & $2.3\times10^{-1}$ & $0.24$ \\
    \phantom{0}600  & $4.2\times10^{-1}$   & $9.0\times10^{-2}$ & $0.21$ \\
    1500            & $1.7\times10^{-1}$   & $3.0\times10^{-2}$ & $0.18$ \\
    3500            & $8.0\times10^{-2}$   & $2.0\times10^{-2}$ & $0.26$ \\
    \bottomrule
  \end{tabular}
\end{table}

\subsection{Architecture and policy-class ablation}
\label{app:ansatz}

The architecture $v(x)=x^\top Sx+\mathrm{MLP}(x)+b$ ($S$, $b$ learned)
\emph{contains} the LQR solution, and a linear-feedback policy class contains
the optimal policy. Table~\ref{tab:ansatz} quantifies the effect of these
structural choices.

\begin{table}[htbp]
  \centering
  \caption{What prior structure is worth: relative $L^2$ error, geometric mean
  over three seeds. The first row is the setting used everywhere else in this
  paper.}
  \label{tab:ansatz}
  \small\setlength{\tabcolsep}{3pt}
  \begin{tabular}{@{}lccc@{}}
    \toprule
    Architecture / policy class & $d=10$ & $d=50$ & gain at $d=50$ \\
    \midrule
    MLP, pointwise greedy (used throughout) & $3.1\times10^{-1}$ & $4.3\times10^{-1}$ & $1$ \\
    MLP, linear-feedback refit & $2.9\times10^{-1}$ & $8.3\times10^{-1}$ & $0.5$ \\
    $x^\top Sx+$MLP, pointwise greedy & $7.8\times10^{-3}$ & $2.5\times10^{-3}$ & $173$ \\
    \bottomrule
  \end{tabular}
\end{table}

The quadratic term improves the error by a factor of $173$ at $d=50$, and its
effect \emph{grows} with dimension ($40$ at $d=10$); the policy restriction
gives no improvement on its own, and it is redundant given the quadratic
term, since the greedy policy of a quadratic value function is already
linear. The
observation transfers as a
design principle: encoding even approximate knowledge of the solution class
outperforms the optimizer variations considered in these ablations. It does
not transfer as a benchmark claim:
a solution inside the hypothesis class stops testing the solver, and the
nonlinear problems of Section~\ref{sec:exp_nonlinear} offer no comparable
structure. The main text therefore reports the first row throughout.

Anchoring $v(0)$ to the analytic $c=\lambda^{-1}\tr(\sigma\sigma^\top P)$
improves the error two- to fivefold, but uses information from the reference
solution during training, so it is used nowhere. That it helps at all reflects the weak determination of the
additive constant: a constant perturbs the residual only through $\lambda$.
An anchor estimated by simulation would be legitimate; future work.

\subsection{Pendulum diagnostics}
\label{app:pendulum_diag}

\paragraph{Budget and granularity controls}
Sweeping the per-evaluation budget from
$N_{\mathrm{eval}}=300$ to $4800$ leaves the peak essentially unchanged ($1.01$ down
to $0.97$): a single policy-evaluation solve already carries the result.
The budget changes what happens afterwards (ten updates cost $0.04$ at
$300$ steps and $0.31$--$0.43$ at every larger budget), while the HJB
residual at the selected iterate \emph{grows} with budget ($6.3$ to
$19.1$), excluding the residual reading. Update granularity does not
substitute either: one hundred small updates reach the same peak near the
eighth and erode to a $0.58$ plateau.

\paragraph{Sampling geometry}
Widening the velocity axis alone recovers part of the Buffered variant's
gain ($0.36\pm0.12$ at matched width); a periodic input embedding, which
removes the $\theta=\pm\pi$ training boundary entirely, recovers only
$0.63\pm0.02$, so the sampling geometry acts through channels the
estimate does not itemize.

\paragraph{Update-side alternatives}
Ces\`aro policy averaging drifts to $0.78$;
growing the training box with $n$ leaves the cycle in place ($0.61$);
the Accepted variant without the anchor (modified policy
iteration) is monotone by construction but
stalls at its first accepted update ($0.982$).

\subsection{Quadrotor remedy comparison}
\label{app:quad_remedies}

The alternatives separate cause from symptom
(Table~\ref{tab:quad_fixes}). Tripling the per-evaluation budget
reaches $0.986$, plus $4\times$ collocation $0.995$, at three times the
cost. Per-coordinate input rescaling ($0.904$) and curvature whitening
($0.841$) address the anisotropy of $V$, which is real (the curvature of
$P$ spans a factor of $216$ across directions against $4$ on the LQR
testbed; $85\%$ of $\|\nabla V\|^2$ lies in a coordinate to which the greedy
map is nearly insensitive) but not binding. Quadrupling the collocation count alone
changes nothing ($0.925$ against $0.924$), so the binding resource is
optimization of the inner problem, not sampling density. The practical
reading: a strongly anisotropic value function needs a per-update budget an
order of magnitude above an isotropic problem of the same dimension, and
the requirement is not visible in the residual: it is detected by the
closed-loop evaluation rather than by a sampled residual diagnostic.

\begin{table}[htbp]
  \centering
  \caption{Quadrotor: three candidate remedies at the default budget, and
  what extra budget achieves instead. Gap closed~\eqref{eq:gap_closed} at the
  monitored peak and at the last of three updates, mean over three seeds
  (two for the budget rows). The cost column counts neural optimizer steps;
  rollout-based variants incur additional model-rollout cost. Only the
  remedy aimed at the sampling distribution
  removes both the shortfall and the degradation.}
  \label{tab:quad_fixes}
  \begin{tabular}{@{}lccc@{}}
    \toprule
    & peak & last & cost \\
    \midrule
    default ($N_{\mathrm{eval}}=4000$) & $0.921$ & $0.858$ & $1\times$ \\
    per-coordinate input scaling       & $0.904$ & $0.810$ & $1\times$ \\
    curvature whitening                & $0.841$ & $0.612$ & $1\times$ \\
    on-policy collocation              & $\mathbf{1.005}$ & $\mathbf{1.005}$ & $1\times$ \\
    \quad without the warm start       & $0.011$ & $0.003$ & $1\times$ \\
    \midrule
    $N_{\mathrm{eval}}=1.2\times10^4$  & $0.986$ & $0.980$ & $3\times$ \\
    \quad $+\,4\times$ collocation     & $0.995$ & $0.995$ & $3\times$ \\
    \bottomrule
  \end{tabular}
\end{table}

\bibliographystyle{elsarticle-harv}
\bibliography{ref.bib}

\end{document}